\documentclass{article}

    \PassOptionsToPackage{numbers, sort&compress}{natbib}
\usepackage[final]{neurips_2023}

\usepackage[utf8]{inputenc} % allow utf-8 input
\usepackage[T1]{fontenc}    % use 8-bit T1 fonts
\usepackage{hyperref}       % hyperlinks
\usepackage{url}            % simple URL typesetting
\usepackage{booktabs}       % professional-quality tables
\usepackage{amsfonts}       % blackboard math symbols
\usepackage{nicefrac}       % compact symbols for 1/2, etc.
\usepackage{microtype}      % microtypography
\usepackage{xcolor}         % colors% 

\usepackage{graphicx}
\usepackage{caption}
\usepackage{subfigure}
\usepackage{makecell}
\usepackage{booktabs}
\usepackage{bm}
\usepackage{xcolor}
\usepackage{float}
\usepackage{multirow}
\usepackage{longtable}
\usepackage{framed}
\usepackage{enumitem}
\usepackage{amsthm,amsmath,amssymb}
\usepackage{pythonhighlight}
\usepackage{appendix}
\usepackage[utf8]{inputenc}
\usepackage{bbding}
\usepackage{tabularx}
\usepackage{dsfont}

\DeclareMathAlphabet{\mathpzc}{OT1}{pzc}{m}{it}

\usepackage{mathtools}
\usepackage{multirow}
\usepackage{algorithm}
\usepackage[noend]{algpseudocode}
\usepackage{color,soul}

\usepackage{xspace}
\def\prbm{\texttt{PRBM}\xspace}
\def\mr{\texttt{MR}\xspace}
\def\mpm{\texttt{DBE}\xspace}

\makeatletter
\DeclareRobustCommand\onedot{\futurelet\@let@token\@onedot}
\def\@onedot{\ifx\@let@token.\else.\null\fi\xspace}
\def\eg{\emph{e.g}\onedot} 
\def\ie{\emph{i.e}\onedot} 
 
\def\etc{\emph{etc}\onedot}

\makeatother

\definecolor{blue_}{RGB}{76, 114, 176}
\definecolor{orange_}{RGB}{221, 132, 82}
\definecolor{upload}{RGB}{47, 85, 151}
\definecolor{download}{RGB}{241, 13, 208}
\definecolor{red_}{RGB}{255, 0, 0}
\definecolor{gray_}{RGB}{127, 127, 127}
\definecolor{green_}{RGB}{1, 128, 0}
\definecolor{sjtured_}{RGB}{192, 0, 0}
\definecolor{sjtugreen_}{RGB}{84, 130, 53}

\newtheorem{definition}{Definition}
\newtheorem{theorem}{Theorem}

\newtheorem{corollary}{Corollary}

\usepackage{pifont}
\usepackage{makecell}

\usepackage[capitalize]{cleveref}
\crefname{section}{Sec.}{Secs.}
\Crefname{section}{Section}{Sections}
\Crefname{table}{Table}{Tables}
\crefname{table}{Tab.}{Tabs.}

\usepackage{wrapfig}

\title{Eliminating Domain Bias for Federated Learning in Representation Space}

\author{
    Jianqing Zhang$^1$, Yang Hua$^2$, Jian Cao$^1$\thanks{Corresponding authors.}, Hao Wang$^3$, \\ \textbf{Tao Song$^1$ Zhengui Xue$^1$, Ruhui Ma$^{1*}$, Haibing Guan$^1$} \\
    $^1$Shanghai Jiao Tong University\quad $^2$Queen's University Belfast\quad $^3$Louisiana State University\\
    {\tt \small \{tsingz, cao-jian, songt333, zhenguixue, ruhuima, hbguan\}@sjtu.edu.cn} \\
    {\tt \small Y.Hua@qub.ac.uk, haowang@lsu.edu}
}

\begin{document}

\maketitle

\begin{abstract}
    Recently, federated learning (FL) is popular for its privacy-preserving and collaborative learning abilities. However, under statistically heterogeneous scenarios, we observe that biased data domains on clients cause a \textit{representation bias} phenomenon and further degenerate generic representations during local training, \ie, the \textit{representation degeneration} phenomenon. To address these issues, we propose a general framework \textbf{\textit{Domain Bias Eliminator}} (\mpm) for FL. Our theoretical analysis reveals that \mpm can promote bi-directional knowledge transfer between server and client, as it reduces the domain discrepancy between server and client in representation space. Besides, extensive experiments on four datasets show that \mpm can greatly improve existing FL methods in both generalization and personalization abilities. The \mpm-equipped FL method can outperform ten state-of-the-art personalized FL methods by a large margin. Our code is public at \url{https://github.com/TsingZ0/DBE}.
\end{abstract}

\section{Introduction}

\setlength{\abovecaptionskip}{0pt}
\setlength{\belowcaptionskip}{18pt}

As a popular distributed machine learning paradigm with excellent privacy-preserving and collaborative learning abilities, federated learning (FL) trains models among clients with their private data kept locally~\cite{yang2019federated, kairouz2019advances, mcmahan2017communication}. 
Traditional FL (\eg, the famous FedAvg~\cite{mcmahan2017communication}) learns one single global model in an iterative manner by locally training models on clients and aggregating client models on the server. 
However, it suffers an accuracy decrease under statistically heterogeneous scenarios, which are common scenarios in practice~\cite{mcmahan2017communication, t2020personalized, li2021ditto, zhang2022fedala}. 
% From the perspective of domain adaptation~\cite{daume2007frustratingly, lee2021dranet, ben2006analysis}, clients own their specific local data domains while the server considers a \textit{virtual} global data domain including all local data domains~\cite{deng2020adaptive, zhu2021data}. In each iteration, the \textbf{\textit{global-to-local knowledge transfer}} happens after clients receive the global model, and the \textbf{\textit{local-to-global knowledge transfer}} also happens after the server receives client models. 

\begin{wrapfigure}{r}{0.5\textwidth}
\vspace{-27pt}
	\centering
	\subfigure[Before.]{\includegraphics[width=0.32\linewidth]{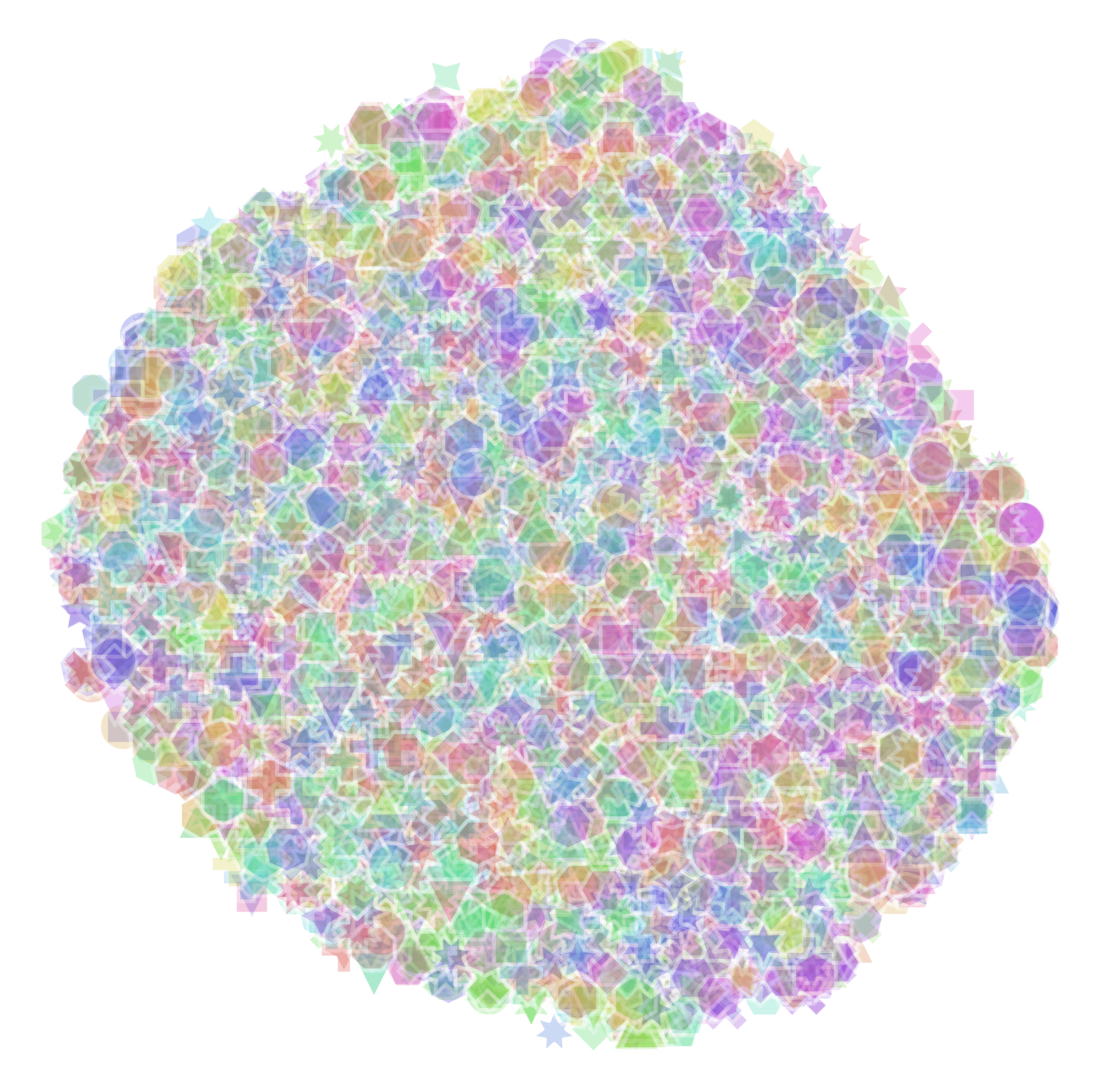}\label{fig:before}}
    \hfill
	\subfigure[After.]{\includegraphics[width=0.32\linewidth]{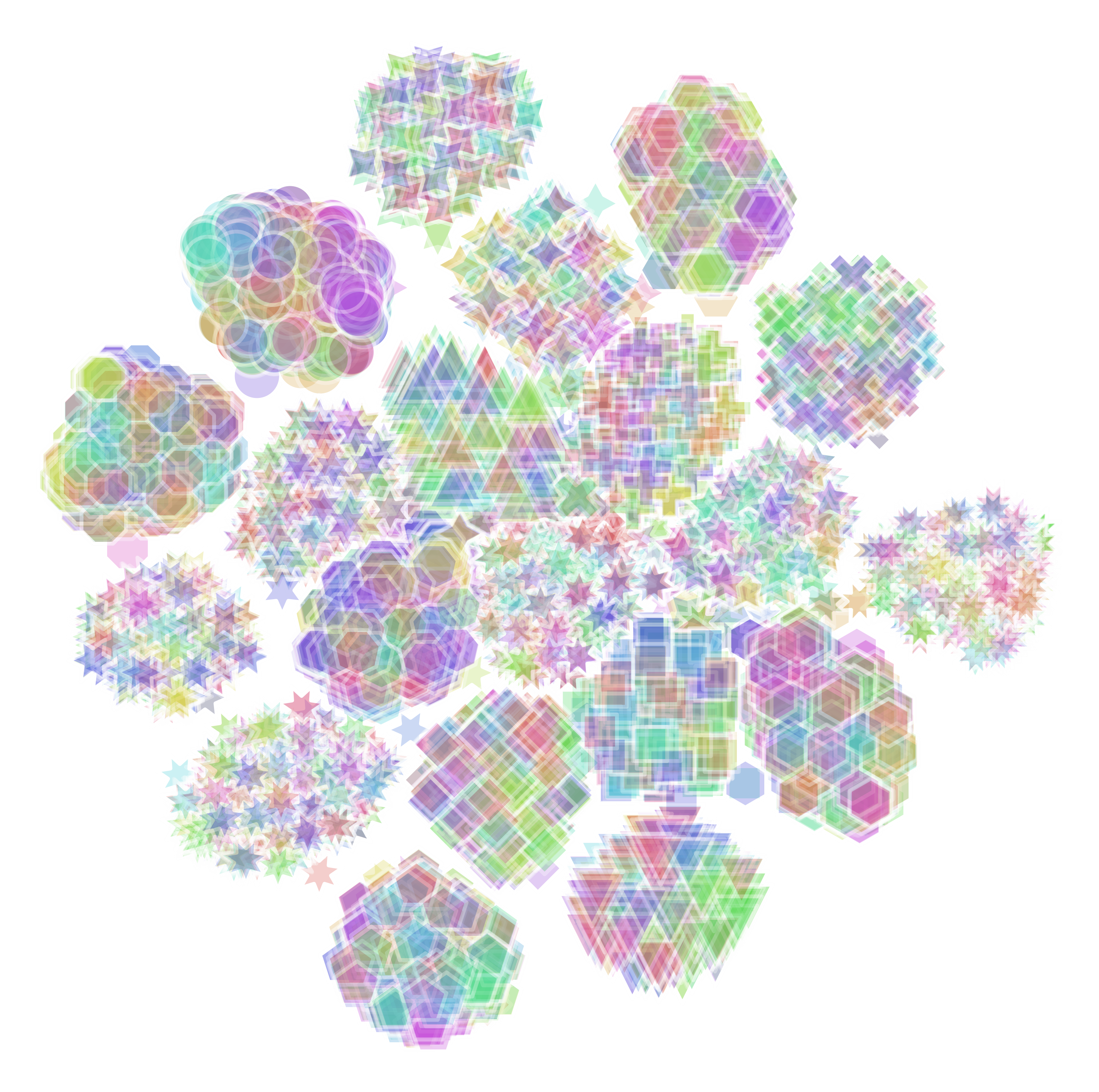}\label{fig:after}}
    \hfill
	\subfigure[MDL change.]{\includegraphics[width=0.33\linewidth]{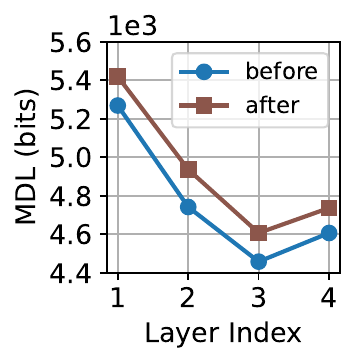}\label{fig:mdl}}
	\caption{t-SNE~\cite{van2008visualizing} visualization and per-layer MDL (bits) for representations before/after local training in FedAvg. We use \textit{color} and \textit{shape} to distinguish \textit{labels} and \textit{clients} respectively for t-SNE. A large MDL means low representation quality. \textit{Best viewed in color and zoom-in.} }
	\label{fig:fedavg}
\vspace{-30pt}
\end{wrapfigure}

% Different from the methods that focus on the federated version of domain adaptation~\cite{zhuang2022federated, yao2022federated, pengfederated, mohri2019agnostic}, we view each iteration in FL itself as domain adaptation. 
% Through model parameters/updates transmitting, clients transfer their learned knowledge to the server, then the server also transfers the aggregated knowledge back to clients in each iteration. 
% However, traditional FL suffers an accuracy decrease in statistically heterogeneous scenarios~\cite{t2020personalized, chen2022personalized, zhang2020personalized}, which are common scenarios in practice~\cite{mcmahan2017communication, li2021ditto, zhang2022fedala}. 

% \begin{figure}[h]
% \vspace{-10pt}
% 	\centering
%     \hfill
% 	\subfigure[Before.]{\includegraphics[width=0.2\linewidth]{fig/study_rep_new_tsne_Tiny-imagenet-0.1_cnn_FedAvg-save_rep_test__01.png}\label{fig:before}}
%     \hfill
% 	\subfigure[After.]{\includegraphics[width=0.2\linewidth]{fig/study_rep_new_tsne_Tiny-imagenet-0.1_cnn_FedAvg-save_rep_test_aft_01.png}\label{fig:after}}
%     \hfill
% 	\subfigure[MDL change.]{\includegraphics[width=0.2\linewidth]{fig/mdl.pdf}\label{fig:mdl}}
%     \hfill
% 	\caption{t-SNE~\cite{van2008visualizing} visualization and per-layer MDL (bits) for representations before/after local training in FedAvg. We use \textit{color} and \textit{shape} to distinguish \textit{labels} and \textit{clients} respectively for t-SNE. Best viewed in color and zoom-in. }
% 	\label{fig:fedavg}
% \vspace{-30pt}
% \end{figure}

Due to statistical heterogeneity, the data domain on each client is biased, which does not contain the data of all labels~\cite{li2021model, kairouz2019advances, yang2019federated, t2020personalized, tan2022towards, zhang2023gpfl}. As the received global model is locally trained on individual clients' biased data domain, we observe that this model extracts biased (\ie, forming client-specific clusters) representations during local training. 
We call this phenomenon ``\textit{representation bias}'' and visualize it in \Cref{fig:fedavg}. 
Meanwhile, by training the received global model with missing labels, 
the generic representation quality over all labels also decreases during local training~\cite{li2021model}. 
Furthermore, we observe that this ``\textit{representation degeneration}'' phenomenon happens at every layer, as shown in \Cref{fig:mdl}. We estimate the representation quality via minimum description length (MDL)~\cite{voita2020information, rogers2021primer, perez2021true}, a metric independent of data and models, measuring the difficulty of classifying target labels according to given representations. 

% which measures the difficulty of classifying target labels given representations and is independent of data and models. 
% Combining our two observations together, we find that the biased data domains on clients lead to representation bias causing representation to degenerate at each layer simultaneously. 

To tackle the statistical heterogeneity, unlike traditional FL that learns a single global model, personalized FL (pFL) comes along by learning personalized models (or modules) for each client besides learning a global model among clients ~\cite{tan2022towards, collins2021exploiting, NEURIPS2020_24389bfe}. Typically, most of the existing pFL methods train a personalized classifier\footnote{A model is split into a feature extractor and a classifier. They are sequentially jointed.} for each client~\cite{chen2021bridging, arivazhagan2019federated, collins2021exploiting, ohfedbabu}, but the feature extractor still extracts all the information from the biased local data domain, leading to representation bias and representation degeneration during local training. 
% Furthermore, they also ignore the significance of generic representation quality. A global model with better generalization ability can transfer more valuable knowledge among clients~\cite{tan2022towards, deng2020adaptive, wang2022generalizing, li2017deeper}. 

% To alleviate the inherent domain shift issue in FL, we propose a framework \textbf{\textit{Domain Bias Eliminator for federated learning}} (\mpm) including two modules introduced as follows. Motivated by the representation bias phenomenon, we detach the biased representation mean from original representations and preserve it in the \textit{\textbf{Personalized Representation Bias Memory}} (\prbm) on each client to let the local feature extractor focus on unbiased representation information. To bring the distributions generated by different local feature extractors close to each other and improve local-to-global knowledge transfer, we devise a \textit{\textbf{Mean Regularization}} (\mr) that explicitly forces local feature extractors to extract representations with a consensual global mean. In this way, we turn one level of representation between the feature extractor and the classifier into two levels of representation with a client-invariant mean and a client-specific mean, respectively. 
% By narrowing the gap between local domains and the \textit{virtual} global domain at both levels of representation, \mpm can promote the bi-directional domain adaptation between server and client with lower generalization bounds. 

To address the representation bias and representation degeneration issues in FL, we propose a general framework \textbf{\textit{Domain Bias Eliminator}} (\mpm) for FL including two modules introduced as follows. Firstly, we detach the representation bias from original representations and preserve it in a \textit{\textbf{Personalized Representation Bias Memory}} (\prbm) on each client. Secondly, we devise a \textit{\textbf{Mean Regularization}} (\mr) that explicitly guides local feature extractors to extract representations with a consensual global mean during local training to let the local feature extractor focus on the remaining unbiased information and improve the generic representation quality. In this way, we turn one level of representation between the feature extractor and the classifier on each client into two levels of representation with a client-specific bias and a client-invariant mean, respectively. Thus, we can eliminate the \textit{conflict} of extracting representations with client-specific biases for clients' requirements while extracting representations with client-invariant features for the server's requirements in the same representation space. 
% Since the global model is generated by aggregating client models on the server, we assume that the server considers a \textit{virtual} global data domain including all local data domains, following prior arts~\cite{deng2020adaptive, zhu2021data}. 
Our theoretical analysis shows that \mpm can promote the bi-directional knowledge transfer between server and client with lower generalization bounds.

We conduct extensive experiments in computer vision (CV) and natural language processing (NLP) fields on various aspects to study the characteristics and effectiveness of \mpm. In both generalization ability (measured by MDL) and personalization ability (measured by accuracy), \mpm can promote the fundamental FedAvg as well as other representative FL methods. 
Furthermore, we compare the representative FedAvg+\mpm with ten state-of-the-art (SOTA) pFL methods in various scenarios and show its superiority over these pFL methods. 
% Since the shape of the trainable parameters in \prbm equals the representation dimension and \mr does not introduce trainable parameters, \mpm costs negligible effort but brings noticeable improvement.
% We provide the code in the supplementary. 
To sum up, our contributions are:

\begin{itemize}
    \item We observe the representation bias and per-layer representation degeneration phenomena during local training in the representative FL method FedAvg. 
    \item We propose a framework \mpm to memorize representation bias on each client to address the representation bias issue and explicitly guide local feature extractors to generate representations with a universal mean for higher generic representation quality. 
    \item We provide theoretical analysis and derive lower generalization bounds of the global and local feature extractors to show that \mpm can facilitate bi-directional knowledge transfer between server and client in each iteration. 
    \item We show that \mpm can improve other representative traditional FL methods including FedAvg at most \textbf{\textcolor{green_}{-22.35\%}} in MDL (bits) and \textbf{\textcolor{green_}{+32.30}} in accuracy (\%), respectively. Furthermore, FedAvg+\mpm outperforms SOTA pFL methods by up to \textbf{\textcolor{green_}{+11.36}} in accuracy (\%). 
\end{itemize}

% $\ \bullet \ $ We observe the representation bias and per-layer representation degeneration phenomena during local training in FedAvg. 

% $\ \bullet \ $ We propose a framework \mpm to memorize representation bias on each client and explicitly guide local feature extractors to generate representations with a universal mean for higher generic representation quality. 

% $\ \bullet \ $ We derive lower generalization bounds of the global and local feature extractors to show that \mpm can facilitate bi-directional knowledge transfer between server and client in each iteration. 

% $\ \bullet \ $ We show that \mpm can improve other representative traditional FL methods including FedAvg at most \textbf{\textcolor{green_}{-22.35\%}} in MDL (bits) and \textbf{\textcolor{green_}{+32.30}} in accuracy (\%), respectively. Furthermore, FedAvg+\mpm outperforms SOTA pFL methods by up to \textbf{\textcolor{green_}{+11.36}} in accuracy (\%). 

\section{Related Work}

Traditional FL methods that focus on improving accuracy under statistically heterogeneous scenarios based on FedAvg including four categories: update-correction-based FL~\cite{karimireddy2020scaffold, niu2022federated, gao2022feddc}, regularization-based FL~\cite{MLSYS2020_38af8613, acarfederated, kim2022multi, cheng2022differentially}, model-split-based FL~\cite{li2021model, jiang2022harmofl}, and knowledge-distillation-based FL~\cite{zhu2021data, zhang2022fine, gong2022preserving, huang2022learn}. For pFL methods, we consider four categories: meta-learning-based pFL~\cite{NEURIPS2020_24389bfe, chen2018federated}, regularization-based pFL~\cite{t2020personalized, li2021ditto}, personalized-aggregation-based pFL~\cite{deng2020adaptive, ijcai2022p301, zhang2020personalized, zhang2023fedala}, and model-split-based pFL~\cite{arivazhagan2019federated, collins2021exploiting, chen2021bridging, ohfedbabu, Zhang2023fedcp}. 
% Domain adaptation is attracting the increasing attention of researchers in FL~\cite{zhuang2022federated, yao2022federated, zeng2022gradient}. However, most of them consider the federated version of domain adaptation (\eg, generalizing to unlabeled data on the server~\cite{pengfederated, sun2023feature, mohri2019agnostic}) but neglect that domain adaptation inherently exists in FL in each iteration. 
Due to limited space, we only introduce the FL methods that are close to ours and leave the \textit{extended version of this section} to \Cref{sec:related}. 

\noindent\textbf{Traditional FL methods.\ \ } MOON~\cite{li2021model} utilizes contrastive learning to correct the local training of each client, but this input-wise contrastive learning still relies on the biased local data domain, so it still suffers from representation skew. 
Although FedGen~\cite{zhu2021data} learns a shared generator on the server and reduces the heterogeneity among clients with the generated representations through knowledge distillation, it only considers the local-to-global knowledge for the single global model learning. 
On the other hand, FedGen additionally brings non-negligible communication and computation overhead for learning and transmitting the generator. 

\noindent\textbf{pFL methods.\ \ } FedPer~\cite{arivazhagan2019federated} and FedRep~\cite{collins2021exploiting} keep the classifier locally, but the feature extractor still learns biased features without explicit guidance. Besides, their local feature extractors are trained to cater to personalized classifiers thus losing generality. 
FedRoD~\cite{chen2021bridging} reduces the discrepancy of local training tasks among clients by using a balanced softmax (BSM) loss function~\cite{ren2020balanced}, but the BSM is useless for missing labels on each client while label missing is a common situation in statistically heterogeneous scenarios~\cite{NEURIPS2020_18df51b9, zhang2022fedala, zhang2020personalized}. Moreover, the uniform label distribution modified by the BSM cannot reflect the original distribution. Differently, FedBABU~\cite{ohfedbabu} trains a global feature extractor with a naive and frozen classifier, then it fine-tunes the classifier for each client to finally obtain personalized models. However, the post-FL fine-tuning study is beyond the FL scope, as almost all the FL methods have multiple fine-tuning variants, \eg, fine-tuning the whole model or only a part of the model. Like FedAvg, FedBABU still locally extracts biased features during the FL process.

\section{Notations and Preliminaries}

\subsection{Notations}

In this work, we discuss the statistically heterogeneous scenario in typical multi-class classification tasks for FL, where $N$ clients share the same model structure. Here, we denote notations following FedGen~\cite{zhu2021data} and FedRep~\cite{collins2021exploiting}. The client $i, i \in [N]$, has its own private data domain $\mathcal{D}_i$, where the data are sampled from $\mathcal{D}_i$. All the clients collaborate to train a global model $g$ parameterized by ${\bm \theta}$ without sharing their private local data. 

Since we focus on representation learning in FL, we regard $g$ as the sequential combination of a feature extractor $f$ that maps from the input space $\mathcal{X}$ to a representation space $\mathcal{Z}$, \ie,  $f: \mathcal{X} \mapsto \mathcal{Z}$ parameterized by ${\bm \theta}^f$ and a classifier $h$ that maps from the representation space to the output space $\triangle^\mathcal{Y}$, \ie, $h: \mathcal{Z} \mapsto \triangle^\mathcal{Y}$ parameterized by ${\bm \theta}^h$. Formally, we have $g := h \circ f$, ${\bm \theta} := [{\bm \theta}^f; {\bm \theta}^h]$, $\mathcal{X} \subset \mathbb{R}^D$ and $\mathcal{Z} \subset \mathbb{R}^K$. $\triangle^\mathcal{Y}$ is the simplex over label space $\mathcal{Y} \subset \mathbb{R}$. With any input ${\bm x} \in \mathcal{X}$, we obtain the feature representation by ${\bm z} = f({\bm x}; {\bm \theta}^f) \in \mathcal{Z}$. 
% Without loss of generality, we define the loss (error) of a model parameterized by ${\bm \theta}$ over a data domain $\mathcal{D}$ as 
% \begin{equation}
%     \mathcal{L}_{\mathcal{D}} ({\bm \theta}) := \mathbb{E}_{({\bm x}, y) \sim \mathcal{D}}[\ell(g({\bm x}; {\bm \theta}), y)] = \mathbb{E}_{({\bm x}, y) \sim \mathcal{D}}[\ell(h(f({\bm x}; {\bm \theta}^f); {\bm \theta}^h), y)] \label{eq:L},
% \end{equation}
% where $\ell: \triangle^\mathcal{Y} \times \mathcal{Y} \mapsto \mathbb{R}$ is a non-negative and convex loss function. 

\subsection{Traditional Federated Learning}

With the collaboration of $N$ clients, the objective of traditional FL, \eg, FedAvg~\cite{mcmahan2017communication}, is to iteratively learn a global model that minimizes its loss on each local data domain:
\begin{equation}
    \min_{{\bm \theta}} \ \mathbb{E}_{i \in [N]}[\mathcal{L}_{\mathcal{D}_i}({\bm \theta})] \label{eq:tFL},
\end{equation}
% \vspace{-10pt}
\begin{equation}
    \mathcal{L}_{\mathcal{D}_i} ({\bm \theta}) := \mathbb{E}_{({\bm x}_i, y_i) \sim \mathcal{D}_i}[\ell(g({\bm x}_i; {\bm \theta}), y_i)] = \mathbb{E}_{({\bm x}_i, y_i) \sim \mathcal{D}_i}[\ell(h(f({\bm x}_i; {\bm \theta}^f); {\bm \theta}^h), y_i)] \label{eq:L},
\end{equation}
where $\ell: \triangle^\mathcal{Y} \times \mathcal{Y} \mapsto \mathbb{R}$ is a non-negative and convex loss function. 
Following FedGen, we assume that all clients share an identical loss function $\ell$ and a \textit{virtual} global data domain $\mathcal{D}$, which is the union of all local domains: $\mathcal{D} := \bigcup^{N}_{i=1} \mathcal{D}_i$. In practice, traditional FL methods~\cite{mcmahan2017communication, MLSYS2020_38af8613, li2021model} optimize \cref{eq:tFL} by $\min_{{\bm \theta}} \sum^{N}_{i=1} \frac{n_i}{n}
 \mathcal{L}_{\hat{\mathcal{D}}_i}({\bm \theta})$, where $\hat{\mathcal{D}}_i$ is an observable dataset, $n_i = |\hat{\mathcal{D}}_i|$ is its size, and $n = \sum^{N}_{i=1} n_i$. 

In each communication iteration, clients conduct local updates on their private data to train the global model ${\bm \theta}$ by minimizing their local loss. Formally, for client $i$, the objective during local training is
$\min_{{\bm \theta}} \ \mathcal{L}_{\mathcal{D}_i}({\bm \theta}).$
The empirical version of $\mathcal{L}_{\mathcal{D}_i}({\bm \theta})$ is $\mathcal{L}_{\hat{\mathcal{D}}_i}({\bm \theta}) := \frac{1}{n_i} \sum^{n_i}_{j=1} \ell(h(f({\bm x}_{ij}; {\bm \theta}^f); {\bm \theta}^h), y_{ij}), $
which is optimized by stochastic gradient descent (SGD)~\cite{zhang2015deep, mcmahan2017communication} in FedAvg. 

\section{Method}

\subsection{Problem Statement}

pFL iteratively learns a personalized model or module for each client with the assistance of the global model parameters from the server. Our objective is (with a slight reuse of the notation $\mathcal{L}_{\mathcal{D}_i}$)
\begin{equation}
    \min_{{\bm \theta}_1, \ldots, {\bm \theta}_N} \ \mathbb{E}_{i \in [N]}[\mathcal{L}_{\mathcal{D}_i}({\bm \theta}_i)] \label{eq:pFL}, 
\end{equation}
where ${\bm \theta}_i$ is a model consisting of global and personalized modules. The global modules are locally trained on clients and shared with the server for aggregation like traditional FL, but the personalized modules are preserved locally on clients. Following traditional FL, we empirically optimize \cref{eq:pFL} by $\min_{{\bm \theta}_1, \ldots, {\bm \theta}_N} \sum^{N}_{i=1} \frac{n_i}{n}
 \mathcal{L}_{\hat{\mathcal{D}}_i}({\bm \theta}_i)$. 

\subsection{Personalized Representation Bias Memory (\prbm)}

\begin{wrapfigure}{r}{0.5\textwidth}
\vspace{-15pt}
	\centering
	\subfigure[Local model (original).]{\includegraphics[width=\linewidth]{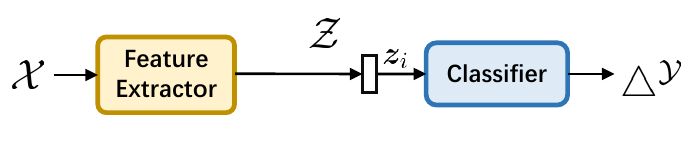}\label{fig:l0}}
    \hfill
	\subfigure[Local model (ours).]{\includegraphics[width=\linewidth]{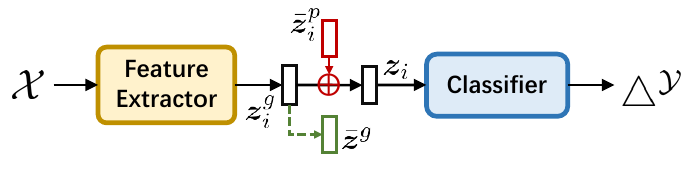}\label{fig:l1}}
	\caption{The illustration of the local model. We emphasize the parts that correspond to \textcolor{sjtured_}{\prbm} and \textcolor{sjtugreen_}{\mr} with \textcolor{sjtured_}{red} and \textcolor{sjtugreen_}{green}, respectively. }
	\label{fig:lm}
\vspace{-15pt}
\end{wrapfigure}

Due to the existence of statistical heterogeneity in FL, the local feature extractor intends to learn biased representations after being overwritten by the received global model parameters. To detach and store the representation bias locally, we propose a personalized module \prbm that memorizes representation bias for client $i$. Originally, the feature representation ${\bm z}_i \in \mathbb{R}^K$ is directly fed into the predictor in \cref{eq:L}. Instead, we consider ${\bm z}_i$ as the combination of a global ${\bm z}^g_i \in \mathbb{R}^K$ and a personalized $\bar{\bm z}^p_i \in \mathbb{R}^K$, \ie, 
\begin{equation}
    {\bm z}_i := {\bm z}^g_i + \bar{\bm z}^p_i. \label{eq:zgi}
\end{equation}
We let the feature extractor output ${\bm z}^g_i$ instead of the original ${\bm z}_i$, \ie, ${\bm z}^g_i := f({\bm x}_i; {\bm \theta}^f)$ and keep the trainable vector $\bar{\bm z}^p_i$ locally. $\bar{\bm z}^p_i$ is specific among clients but identical for all the local data on one client, so it memorizes client-specific mean. The original feature extractor is trained to capture the biased features for ${\bm z}_i$. Instead, with the personalized mean stored in $\bar{\bm z}^p_i$, the feature extractor turns to capture ${\bm z}^g_i$ with less biased feature information. We illustrate the difference between the original approach and our method in \Cref{fig:lm} (\textcolor{sjtured_}{\prbm}). Then, we define the local objective as $\min_{{\bm \theta}_i} \ \mathcal{L}_{\mathcal{D}_i}({\bm \theta}_i)$,
where ${\bm \theta}_i := [{\bm \theta}^f; \bar{\bm z}^p_i; {\bm \theta}^h]$, 
\begin{equation}
    \mathcal{L}_{\mathcal{D}_i}({\bm \theta}_i) := \mathbb{E}_{({\bm x}_i, y_i) \sim \mathcal{D}_i}[\ell(h(f({\bm x}_i; {\bm \theta}^f) + \bar{\bm z}^p_i; {\bm \theta}^h), y_i)] \label{eq:local_p}.
\end{equation}

From the view of transformation, we rewrite \cref{eq:local_p} to
\begin{equation}
    \mathcal{L}_{\mathcal{D}_i}({\bm \theta}_i) := \mathbb{E}_{({\bm x}_i, y_i) \sim \mathcal{D}_i}[\ell(h(\prbm(f({\bm x}_i; {\bm \theta}^f); \bar{\bm z}^p_i); {\bm \theta}^h), y_i)] \label{eq:local_trans},
\end{equation}
where $\prbm: \mathcal{Z} \mapsto \mathcal{Z}$ a personalized \textit{translation} transformation~\cite{xue2020affine} parameterized by $\bar{\bm z}^p_i$. Formally, $\prbm({\bm z}^g_i; \bar{\bm z}^p_i) = {\bm z}^g_i + \bar{\bm z}^p_i, \forall \ {\bm z}^g_i \in \mathcal{Z}$. With $\prbm$, we create an additional level of representation ${\bm z}^g_i$ besides the original level of representation ${\bm z}_i$. We call ${\bm z}^g_i$ and ${\bm z}_i$ as the \textit{first and second levels of representation}, respectively. For the original local model (\Cref{fig:l0}), we have ${\bm z}^g_i \equiv {\bm z}_i$.

\subsection{Mean Regularization (\mr)}

% The closer the two distributions in the representation space are, the easier the knowledge transfer between two domains~\cite{ben2006analysis, zhao2019learning}. 
Without explicit guidance, it is hard for the feature extractor and the \textit{trainable} \textcolor{sjtured_}{\prbm} to distinguish between unbiased and biased information in representations automatically. 
Therefore, to let the feature extractor focus on the unbiased information and further separate ${\bm z}^g_i$ and $\bar{\bm z}^p_i$, we propose an \mr that explicitly guides the local feature extractor to generate ${\bm z}^g_i$ with the help of a client-invariant mean, which is opposite to the client-specific mean memorized in $\bar{\bm z}^p_i$, as shown in \Cref{fig:lm} (\textcolor{sjtugreen_}{\mr}). Specifically, we regularize the mean of ${\bm z}^g_i$ to the consensual global mean $\bar{{\bm z}}^g$ at each feature dimension independently. We then modify \cref{eq:local_trans} as 
\begin{equation}
    \mathcal{L}_{\mathcal{D}_i}({\bm \theta}_i) := \mathbb{E}_{({\bm x}_i, y_i) \sim \mathcal{D}_i}[\ell(h(\prbm(f({\bm x}_i; {\bm \theta}^f); \bar{\bm z}^p_i); {\bm \theta}^h), y_i)] + \kappa \cdot \mr(\bar{{\bm z}}^g_i, \bar{{\bm z}}^g),
\end{equation}
where $\bar{{\bm z}}^g_i = \mathbb{E}_{({\bm x}_i, y_i) \sim \mathcal{D}_i}[f({\bm x}_i; {\bm \theta}^f)]$. We obtain the consensus $\bar{{\bm z}}^g = \sum^{N}_{i=1} \bar{{\bm z}}^g_i$ \textit{during the initialization period before FL (see \Cref{algo})}. We measure the distance of $\bar{{\bm z}}^g_i$ and $\bar{{\bm z}}^g$ by mean squared error (MSE)~\cite{tuchler2002minimum}, and $\kappa$ is a hyperparameter to control the importance of \mr for different tasks. Empirically,
\begin{equation}
    \mathcal{L}_{\hat{\mathcal{D}}_i}({\bm \theta}_i) := \frac{1}{n_i} \sum^{n_i}_{j=1} \ell(h(\prbm(f({\bm x}_{ij}; {\bm \theta}^f); \bar{\bm z}^p_i); {\bm \theta}^h), y_{ij}) + \kappa \cdot \mr(\frac{1}{n_i} \sum^{n_i}_{j=1} f({\bm x}_{ij}; {\bm \theta}^f), \bar{{\bm z}}^g), \label{eq:mr}
\end{equation}
which is also optimized by SGD following FedAvg. 

In \cref{eq:mr}, the value of the \mr term is obtained after calculating the empirical version of $\bar{{\bm z}}^g_i$: $\hat{\bar{{\bm z}}}^g_i = \frac{1}{n_i} \sum^{n_i}_{j=1} f({\bm x}_{ij}; {\bm \theta}^f)$ over the entire local data, but the loss computing in SGD cannot see all the local data during one forward pass in one batch. In practice, inspired by the widely-used moving average~\cite{zhang2015deep, liconvergence} in approximating statistics over data among batches, in each batch, we obtain 
\begin{equation}
    \hat{\bar{{\bm z}}}^g_i = (1 - \mu) \cdot \hat{\bar{{\bm z}}}^g_{i, old} + \mu \cdot \hat{\bar{{\bm z}}}^g_{i, new}, \label{eq:moving} 
\end{equation}
where $\hat{\bar{{\bm z}}}^g_{i, old}$ and $\hat{\bar{{\bm z}}}^g_{i, new}$ are computed in the previous batch and current batch, respectively. $\mu$ is a hyperparameter called momentum that controls the importance of the current batch. The feature extractor is updated continuously during local training but discontinuously between adjacent two iterations due to server aggregation. Thus, we only calculate $\hat{\bar{{\bm z}}}^g_i$ via \cref{eq:moving} during local training and recalculate it in a new iteration without using its historical records. We consider the representative FedAvg+\mpm as an example and show the entire learning process in \Cref{algo}. 

\begin{algorithm}[ht]
	\caption{The Learning Process in FedAvg+\mpm}
	\begin{algorithmic}[1]
		\Require 
		$N$ clients with their local data; 
		initial parameters ${\bm \theta}^{f, 0}$ and ${\bm \theta}^{h, 0}$; 
		$\eta$: local learning rate; 
		$\kappa$ and $\mu$: hyperparameters;
		$\rho$: client joining ratio; 
		$E$: local epochs; 
		$T$: total communication iterations. 
		\Ensure 
		Global model parameters $\{{\bm \theta}^f, {\bm \theta}^h\}$ and personalized model parameters $\{\bar{\bm z}^p_1, \ldots, \bar{\bm z}^p_N\}$.
        \Statex \Comment{\textbf{\textit{Initialization Period}}}
        \State Server sends $\{{\bm \theta}^{f, 0}, {\bm \theta}^{h, 0}\}$ to all clients to initialize their local models. 
		\State $N$ clients train their local models \textit{without \mpm} for one epoch and collect client-specific mean $\{\bar{{\bm z}}^g_1, \ldots, \bar{{\bm z}}^g_N\}$ over their data domain. 
		\State Server generates a consensual global mean $\bar{{\bm z}}^g$ through weighted averaging: $\bar{{\bm z}}^g = \sum^{N}_{i=1} \frac{n_i}{n} \bar{{\bm z}}^g_i$. 
		\State Client $i$ initializes $\bar{\bm z}^{p, 0}_i$, $\forall i \in [N]$. 
        \Statex \Comment{\textbf{\textit{Federated Learning Period}}}
		\For{communication iteration $t=1, \ldots, T$}
		    \State Server samples a client subset $\mathcal{I}^t$ based on $\rho$.
		    \State Server sends $\{{\bm \theta}^{f, t-1}, {\bm \theta}^{h, t-1}\}$ to each client in $\mathcal{I}^t$.
		    \For{Client $i \in \mathcal{I}^t$ in parallel}
		        \State Initialize $f$ and $h$ with ${\bm \theta}^{f, t-1}$ and ${\bm \theta}^{h, t-1}$, respectively. 
                \State Obtain $\{{\bm \theta}^{f, t}_i, \bar{\bm z}^{p, t}_i, {\bm \theta}^{h, t}_i\}$ using SGD for $\min_{{\bm \theta}_i} \ \mathcal{L}_{\hat{\mathcal{D}}_i}({\bm \theta}_i)$ with $\eta$, $\kappa$ and $\mu$ for $E$ epochs. 
		        \State Upload $\{{\bm \theta}^{f, t}_i, {\bm \theta}^{h, t}_i\}$ to the server. 
		    \EndFor
		    \State Server calculates $n^t = \sum_{i \in
		    \mathcal{I}^t} n_i$ and obtains 
		        \State \qquad ${\bm \theta}^{f, t} = \sum_{i \in \mathcal{I}^t} \frac{n_i}{n^t} {\bm \theta}^{f, t}_i$; 
		        \State \qquad ${\bm \theta}^{h, t} = \sum_{i \in \mathcal{I}^t} \frac{n_i}{n^t} {\bm \theta}^{h, t}_i$.
		\EndFor
		\\
		\Return $\{{\bm \theta}^{f, T}, {\bm \theta}^{h, T}\}$ and $\{\bar{\bm z}^{p, T}_1, \ldots, \bar{\bm z}^{p, T}_N\}$
	\end{algorithmic}
	\label{algo}
\end{algorithm}

\subsection{Improved Bi-directional Knowledge Transfer}

In the FL field, prior methods draw a connection from FL to domain adaptation for theoretical analysis and consider a binary classification problem~\cite{deng2020adaptive, zhu2021data, tan2022towards, mansour2020three}. 
The traditional FL methods, which focus on enhancing the performance of a global model, regard local domains $\mathcal{D}_i, i\in [N]$ and the virtual global domain $\mathcal{D}$ as the source domain and the target domain, respectively~\cite{zhu2021data}, which is called local-to-global knowledge transfer in this paper. In contrast, pFL methods that focus on improving the performance of personalized models regard $\mathcal{D}$ and $\mathcal{D}_i, i\in [N]$ as the source domain and the target domain, respectively~\cite{deng2020adaptive, tan2022towards, mansour2020three}. We call this kind of adaptation as global-to-local knowledge transfer. 
The local-to-global knowledge transfer happens on the server while the global-to-local one occurs on the client. 
Please refer to \Cref{sec:theo} for details and proofs. 

\subsubsection{Local-To-Global Knowledge Transfer}

Here, we consider the transfer after the server receives a client model. 
We guide the feature extractor to learn representations with a global mean and gradually narrow the gap between the local domain and global domain at the first level of representation (\ie, ${\bm z}^g_i$) to improve knowledge transfer:

\begin{corollary}
    Consider a local data domain $\mathcal{D}_i$ and a virtual global data domain $\mathcal{D}$ for client $i$ and the server, respectively. Let $\mathcal{D}_i = \langle \mathcal{U}_i, c^* \rangle$ and $\mathcal{D} = \langle \mathcal{U}, c^* \rangle$, where $c^*: \mathcal{X} \mapsto \mathcal{Y}$ is a ground-truth labeling function. Let $\mathcal{H}$ be a hypothesis space of VC dimension $d$ and $h: \mathcal{Z} \mapsto \mathcal{Y}, \forall \ h \in \mathcal{H}$. When using \mpm, given a feature extraction function $\mathcal{F}^g: \mathcal{X} \mapsto \mathcal{Z}$ that shared between $\mathcal{D}_i$ and $\mathcal{D}$, a random labeled sample of size $m$ generated by applying $\mathcal{F}^g$ to a random sample from $\mathcal{U}_i$ labeled according to $c^*$, then for every $h^g\in \mathcal{H}$, with probability at least $1-\delta$: 
    $$ 
    \mathcal{L}_{\mathcal{D}}(h^g) \le \mathcal{L}_{\hat{\mathcal{D}}_i}(h^g) + \sqrt{\frac{4}{m} (d\log\frac{2em}{d} + \log\frac{4}{\delta})} + d_{\mathcal{H}}(\tilde{\mathcal{U}}^g_i, \tilde{\mathcal{U}}^g) + \lambda_i, 
    $$
    where $\mathcal{L}_{\hat{\mathcal{D}}_i}$ is the empirical loss on $\mathcal{D}_i$, $e$ is the base of the natural logarithm, and $d_{\mathcal{H}}(\cdot, \cdot)$ is the $\mathcal{H}$-divergence between two distributions. $\lambda_i := \min_{h^g} \mathcal{L}_{\mathcal{D}}(h^g) + \mathcal{L}_{\mathcal{D}_i}(h^g)$, $\tilde{\mathcal{U}}^g_i \subseteq \mathcal{Z}$, $\tilde{\mathcal{U}}^g \subseteq \mathcal{Z}$, and $d_{\mathcal{H}}(\tilde{\mathcal{U}}^g_i, \tilde{\mathcal{U}}^g) \le d_{\mathcal{H}}(\tilde{\mathcal{U}}_i, \tilde{\mathcal{U}})$. $\tilde{\mathcal{U}}^g_i$ and $\tilde{\mathcal{U}}^g$ are the induced distributions of $\mathcal{U}_i$ and $\mathcal{U}$ under $\mathcal{F}^g$, respectively. $\tilde{\mathcal{U}}_i$ and $\tilde{\mathcal{U}}$ are the induced distributions of $\mathcal{U}_i$ and $\mathcal{U}$ under $\mathcal{F}$, respectively. $\mathcal{F}$ is the feature extraction function in the original FedAvg without \mpm. \label{corollary:mr}
\end{corollary}

As shown in \Cref{fig:lm}, given any ${\bm x}_i$ on client $i$, one can obtain ${\bm z}_i$ via $\mathcal{F}$ in original FedAvg or obtain ${\bm z}^g_i$ via $\mathcal{F}^g$ in FedAvg+\mpm. 
With $d_{\mathcal{H}}(\tilde{\mathcal{U}}^g_i, \tilde{\mathcal{U}}^g) \le d_{\mathcal{H}}(\tilde{\mathcal{U}}_i, \tilde{\mathcal{U}})$ holds, we can achieve a lower generalization bound in local-to-global knowledge transfer than traditional FL, thus training a better global feature extractor to produce representations with higher quality over all labels. A small gap between the local domain and global domain in $\mathcal{Z}$ promotes the knowledge transfer from clients to the server~\cite{zhang2012generalization, zhao2018adversarial, zhang2019bridging}. 

\subsubsection{Global-To-Local Knowledge Transfer}

The global-to-local knowledge transfer focuses on the assistance role of the global model parameters for facilitating local training, \ie, the transfer ability from $\mathcal{D}$ to $\mathcal{D}_i$. After the client receives the global model and equips it with \prbm, for the second level of representation (\ie, ${\bm z}_i$), we have

\begin{corollary}
    Let $\mathcal{D}_i$, $\mathcal{D}$, $\mathcal{F}^g$, and $\lambda_i$ defined as in \cref{corollary:mr}. Given a translation transformation function $\prbm: \mathcal{Z} \mapsto \mathcal{Z}$ that shared between $\mathcal{D}_i$ and \textit{virtual} $\mathcal{D}$, a random labeled sample of size $m$ generated by applying $\mathcal{F}'$ to a random sample from $\mathcal{U}_i$ labeled according to $c^*$, $\mathcal{F}' = \prbm \circ \mathcal{F}^g : \mathcal{X} \mapsto \mathcal{Z}$, then for every $h'\in \mathcal{H}$, with probability at least $1-\delta$: 
    $$ 
    \mathcal{L}_{\mathcal{D}_i}(h') \le \mathcal{L}_{\hat{\mathcal{D}}}(h') + \sqrt{\frac{4}{m} (d\log\frac{2em}{d} + \log\frac{4}{\delta})} + d_{\mathcal{H}}(\tilde{\mathcal{U}}', \tilde{\mathcal{U}}'_i) + \lambda_i, 
    $$
    where $d_{\mathcal{H}}(\tilde{\mathcal{U}}', \tilde{\mathcal{U}}'_i) = d_{\mathcal{H}}(\tilde{\mathcal{U}}^g, \tilde{\mathcal{U}}^g_i) \le d_{\mathcal{H}}(\tilde{\mathcal{U}}, \tilde{\mathcal{U}}_i) = d_{\mathcal{H}}(\tilde{\mathcal{U}_i}, \tilde{\mathcal{U}})$.  $\tilde{\mathcal{U}}'$ and $\tilde{\mathcal{U}}'_i$ are the induced distributions of $\mathcal{U}$ and $\mathcal{U}_i$ under $\mathcal{F}'$, respectively.  \label{corollary:pm}
\end{corollary}

Given ${\bm x}_i$ on client $i$, we can obtain ${\bm z}_i$ via $\mathcal{F}'$ in FedAvg+\mpm. $h^g = h' \circ \prbm$, so \prbm does not influence the value of $d_{\mathcal{H}}(\cdot, \cdot)$ for the pair of $h^g$ and $h'$ (see \Cref{sec:corollary2}), then we have $d_{\mathcal{H}}(\tilde{\mathcal{U}}', \tilde{\mathcal{U}}'_i) = d_{\mathcal{H}}(\tilde{\mathcal{U}}^g, \tilde{\mathcal{U}}^g_i)$. The inequality $d_{\mathcal{H}}(\tilde{\mathcal{U}}', \tilde{\mathcal{U}}'_i) \le d_{\mathcal{H}}(\tilde{\mathcal{U}}, \tilde{\mathcal{U}}_i)$ shows that the information aggregated on the server can be more easily transferred to clients with our proposed \mpm than FedAvg. We train \prbm on the local loss and preserve it locally, so the local feature extractors can generate representations suitable for clients' personalized tasks. According to \cref{corollary:mr} and \cref{corollary:pm}, adding \mpm facilitates the bi-directional knowledge transfer in each iteration, gradually promoting global and local model learning as the number of iterations increases. 

\subsection{Negligible Additional Communication and Computation Overhead}

% \begin{figure}[h] \vspace{-15pt}
% 	\centering
% 	\subfigure[Local model (\textcolor{red_}{original}).]{\includegraphics[width=0.49\linewidth]{fig/local model0.pdf}\label{fig:l0}}
%     \hfill
% 	\subfigure[Local model (\textcolor{red_}{ours}).]{\includegraphics[width=0.49\linewidth]{fig/local model1.pdf}\label{fig:l1}}
% 	\caption{Illustration of the local model. We emphasize the parts that correspond to \textcolor{sjtugreen_}{\mr} and \textcolor{sjtured_}{\prbm} with \textcolor{sjtugreen_}{green} and \textcolor{sjtured_}{red}, respectively. }
% 	\label{fig:lm}
% \vspace{-15pt} \end{figure}

% By integrating our proposed \mpm with FedAvg, we illustrate the original local model and our modified local model in \Cref{fig:lm}. 
% We emphasize the parts that correspond to \textcolor{sjtugreen_}{\mr} and \textcolor{sjtured_}{\prbm} with \textcolor{sjtugreen_}{green} and \textcolor{sjtured_}{red}, respectively. 
\mpm only modifies the local training, so the downloading, uploading, and aggregation processes in FedAvg are unaffected. In FedAvg+\mpm, the communication overhead per iteration is the same as FedAvg but requires fewer iterations to converge (see \Cref{sec:addexp}). Moreover, 
\prbm only introduces $K$ additional trainable parameters, and the MSE value in the parameterless \mr is computed for two representations of $K$ dimension. 
$K$ is the representation space dimension, typically a smaller value than the dimension of data inputs or model parameters~\cite{bengio2013representation, zhang2018network}. 
Thus, \mpm introduces no additional communication overhead and negligible computation overhead for local training in any iteration. 

\subsection{Privacy-Preserving Discussion}
Compared to FedAvg, using \mpm requires client $i$ to upload one client-specific mean $\bar{{z}}^g_i$ (one $K$-dimensional vector) to the server \textbf{\textit{only once}} \textit{before FL}, which solely captures the magnitude of the mean value for each feature dimension within the context of the given datasets and models. Thanks to this particular characteristic, as shown in \Cref{sec:privacy}, the performance of FedAvg+\mpm can be minimally affected while enhancing its privacy-preserving capabilities by introducing proper Gaussian noise with a zero mean to $\bar{{z}}^g_i$ during the initialization phase. 
% On the other hand, for each client, FedAvg+\mpm shares \textit{one} $K$-dimensional vector \textit{only once} during the initialization period, while existing methods that share feature information~\cite{xu2022personalized, tan2022fedproto, tan2022federated} upload \textit{multiple} class-wise $K$-dimensional vector in each communication iteration. Therefore, FedAvg+\mpm shares much less information than those methods. 

\section{Experiments}

% \subsection{Setup}

\noindent\textbf{Datasets and models. \ } Following prior FL approaches~\cite{mcmahan2017communication, li2021model, NEURIPS2020_18df51b9, collins2021exploiting, chen2021bridging}, we use four public datasets for classification problems in FL, including three CV datasets: Fashion-MNIST (FMNIST)~\cite{xiao2017fashion}, Cifar100~\cite{krizhevsky2009learning}, and Tiny-ImageNet (100K images with 200 labels)~\cite{chrabaszcz2017downsampled}, as well as one NLP dataset: AG News~\cite{zhang2015character}. For three CV datasets, we adopt the popular 4-layer CNN by default following FedAvg, which contains two convolution layers (denoted by CONV1 and CONV2) and two fully connected layers (denoted by FC1 and FC2). Besides, we also use a larger model ResNet-18~\cite{he2016deep} on Tiny-ImageNet. For AG News, we use the famous text classification model fastText~\cite{joulinetal2017bag}. 

\noindent\textbf{Statistically heterogeneous scenarios. \ } There are two widely used approaches to construct statistically heterogeneous scenarios on public datasets: the pathological setting~\cite{mcmahan2017communication, pmlrv139shamsian21a} and practical setting~\cite{NEURIPS2020_18df51b9, li2021model}. For the pathological setting, disjoint data with 2/10/20 labels for each client are sampled from 10/100/200 labels on FMNIST/Cifar100/Tiny-ImageNet with different data amounts. For the practical setting, we sample data from FMNIST, Cifar100, Tiny-ImageNet, and AG News based on the Dirichlet distribution~\cite{NEURIPS2020_18df51b9} (denoted by $Dir(\beta)$). Specifically, we allocate a $q_{c, i}$ ($q_{c, i} \sim Dir(\beta)$) proportion of samples with label $c$ to client $i$, and we set $\beta=0.1$/$\beta=1$ by default for CV/NLP tasks following previous FL approaches~\cite{NEURIPS2020_564127c0, NEURIPS2020_18df51b9}. 

\noindent\textbf{Implementation Details. \ } Following pFedMe and FedRoD, we have 20 clients and set client participating ratio $\rho=1$ by default unless otherwise stated. We measure the generic representation quality across clients and evaluate the MDL~\cite{voita2020information, rogers2021primer} of representations over all class labels. To simulate the common FL scenario where data only exists on clients, we split the data among each client into two parts: a training set (75\% data) and a test set (25\% data). Following pFedMe, we evaluate pFL methods by averaging the results of personalized models on the test set of each client and evaluate traditional FL methods by averaging the results of the global model on each client. 
% To further evaluate the generalization ability, we perform additional local training for 10 epochs on new clients to obtain the post-FL fine-tuning results. The local training procedure depends on the FL method design. 
Following FedAvg, we set the batch size to 10 and the number of local epochs to 1, so the number of local SGD steps is $\lfloor \frac{n_i}{10} \rfloor$ for client $i$. We run three trials for all methods until empirical convergence on each task and report the mean value. 
For more details and results (\eg, fine-tuning FedAvg on new participants and a real-world application), please refer to \Cref{sec:addexp}. 

\subsection{Experimental Study for Adding \mpm}

\subsubsection{How to Split the Model?}

% Given any input, each layer in a deep neural network (DNN) outputs a feature representation and feeds it into the next layer~\cite{lecun2015deep, bengio2013representation}. 
A model is split into a feature extractor and a classifier, but there are various ways for splitting, as each layer in a deep neural network (DNN) outputs a feature representation and feeds it into the next layer~\cite{lecun2015deep, bengio2013representation}~\cite{lecun2015deep, bengio2013representation}. We focus on inserting \mpm between the feature extractor and the classifier, but which splitting way is the best for \mpm? Here we answer this question by comparing the results regarding MDL and accuracy when the model is split at each layer in the popular 4-layer CNN. We show the MDL of the representation ${\bm z}^g_i$ outputted by the prepositive layer of \mpm (with \underline{underline} here) and show MDL of ${\bm z}_i$ for other layers. 
Low MDL and high accuracy indicate superior generalization ability and superior personalization ability, respectively. 

\begin{table}[h]
  \centering
  \caption{The MDL (bits, $\downarrow$) of layer-wise representations, test accuracy (\%, $\uparrow$), and the number of trainable parameters ($\downarrow$) in \prbm when adding \mpm to FedAvg on Tiny-ImageNet using 4-layer CNN in the practical setting. We also show corresponding results for the close pFL methods. For FedBABU, ``[36.82]'' indicates the test accuracy after post-FL fine-tuning for 10 local epochs. }
  \resizebox{\linewidth}{!}{
    \begin{tabular}{l|cccc|c|c}
    \toprule
    \multirow{2}{*}{\textbf{Metrics}} & \multicolumn{4}{c|}{\textbf{MDL}} & \multirow{2}{*}{\textbf{Accuracy}} & \multirow{2}{*}{\textbf{Param.}} \\
    \cmidrule{2-5}
     & CONV1$\rightarrow$CONV2 & CONV2$\rightarrow$FC1 & FC1$\rightarrow$FC2 & Logits & & \\
    \midrule
    FedPer~\cite{arivazhagan2019federated} & 5143 & 4574 & 3885 & 4169 & 33.84 & --- \\
    FedRep~\cite{collins2021exploiting} & 5102 & 4237 & 3922 & 4244 & 37.27 & --- \\
    FedRoD~\cite{chen2021bridging} & 5063 & 4264 & 3783 & 3820 & 36.43 & --- \\
    FedBABU~\cite{ohfedbabu} & 5083 & 4181 & 3948 & 3849 & 16.86 [36.82] & --- \\
    \midrule
    Original (FedAvg) & 5081 & 4151 & 3844 & 3895 & 19.46 & 0 \\
    CONV1$\rightarrow$\mpm$\rightarrow$CONV2 & \underline{4650} (\textcolor{green_}{-8.48\%}) & 4105 (\textcolor{green_}{-1.11\%}) & 3679 (\textcolor{green_}{-4.29\%}) & 3756 (\textcolor{green_}{-3.57\%}) & 21.81 (\textcolor{green_}{+2.35}) & 28800 \\
    CONV2$\rightarrow$\mpm$\rightarrow$FC1 & \textbf{4348 (\textcolor{green_}{-14.43\%})} & \underline{3716} (\textcolor{green_}{-10.48\%}) & \textbf{3463 (\textcolor{green_}{-9.91\%})} & \textbf{3602 (\textcolor{green_}{-7.52\%})} & \textbf{47.03 (\textcolor{green_}{+27.57})} & 10816 \\
    FC1$\rightarrow$\mpm$\rightarrow$FC2 & 4608 (\textcolor{green_}{-9.31\%}) & \textbf{3689 (\textcolor{green_}{-11.13\%})} & \underline{3625} (\textcolor{green_}{-5.70\%}) & 3688 (\textcolor{green_}{-5.31\%}) & 43.32 (\textcolor{green_}{+23.86}) & 512 \\
    % FC2$\rightarrow$\mpm$\rightarrow$OUTPUT & 5005 (\textcolor{green_}{-1.50\%}) & 3954 (\textcolor{green_}{-4.75\%}) & 3644 (\textcolor{green_}{-5.20\%}) & \underline{4071} (4.52\%) & 34.45 (\textcolor{green_}{+14.99}) & 200 \\
    \bottomrule
    \end{tabular}}
  \label{tab:perlayer_cnn_tiny}
\vspace{-5pt} \end{table}

% A small MDL value means high generic representation quality. As deeper models are better at learning representations~\cite{lecun2015deep, bengio2013representation, he2016deep}, the representations outputted by a deeper layer have higher generic representation quality (lower MDL) in \Cref{tab:perlayer_cnn_tiny}. 
In \Cref{tab:perlayer_cnn_tiny}, the generic representation quality is improved at each layer for all splitting ways, which shows that no matter how the model is split, \mpm can enhance the generalization ability of the global feature extractor. Among these splitting ways, assigning all FC layers to the classifier, \ie, CONV2$\rightarrow$\mpm$\rightarrow$FC1, achieves almost the lowest MDL and highest accuracy. 
% According to \Cref{tab:perlayer_cnn_tiny}, adding \mpm before CONV2 does not bring too much MDL and accuracy improvement, while adding \mpm before FC layers (FC1 and FC2) brings more MDL and accuracy improvement, especially for accuracy. 
% However, the representation dimension in the lower layer is higher (\ie, larger $K$) than the one in the higher layer~\cite{lecun2015deep}. Thus, \mpm introduces more than $21\times$ trainable parameters when we insert it between lower layers. For example, CONV2$\rightarrow$\mpm$\rightarrow$FC1 requires $21.13\times$ trainable parameters for \mpm than FC1$\rightarrow$\mpm$\rightarrow$FC2. 
Meanwhile, FC1$\rightarrow$\mpm$\rightarrow$FC2 can also achieve excellent performance with \textbf{only $4.73\%$} trainable parameters for \mpm. 

Since FedRep, FedRoD, and FedBABU choose the last FC layer as the classifier by default, we follow them for a fair comparison and insert \mpm before the last FC layer (\eg, FC1$\rightarrow$\mpm$\rightarrow$FC2). In \Cref{tab:perlayer_cnn_tiny}, our FC1$\rightarrow$\mpm$\rightarrow$FC2 outperforms FedPer, FedRep, FedRoD, FedBABU, and FedAvg with lower MDL and higher accuracy. Since feature extractors in FedPer and FedRep are locally trained to cater to personalized classifiers, they extract representations with low quality.

\subsubsection{Representation Bias Eliminated for the First Level of Representation}

\begin{figure}[h]
\vspace{-10pt}
	\centering
	\subfigure[FedAvg (B).]{\includegraphics[width=0.16\linewidth]{fig/study_rep_new_tsne_Tiny-imagenet-0.1_cnn_FedAvg-save_rep_test__01.png}\label{fig:fed_before}}
    \hfill
	\subfigure[FedAvg (A).]{\includegraphics[width=0.16\linewidth]{fig/study_rep_new_tsne_Tiny-imagenet-0.1_cnn_FedAvg-save_rep_test_aft_01.png}\label{fig:fed_after}}
	\subfigure[+\mpm (${\bm z}^g_i$, B).]{\includegraphics[width=0.16\linewidth]{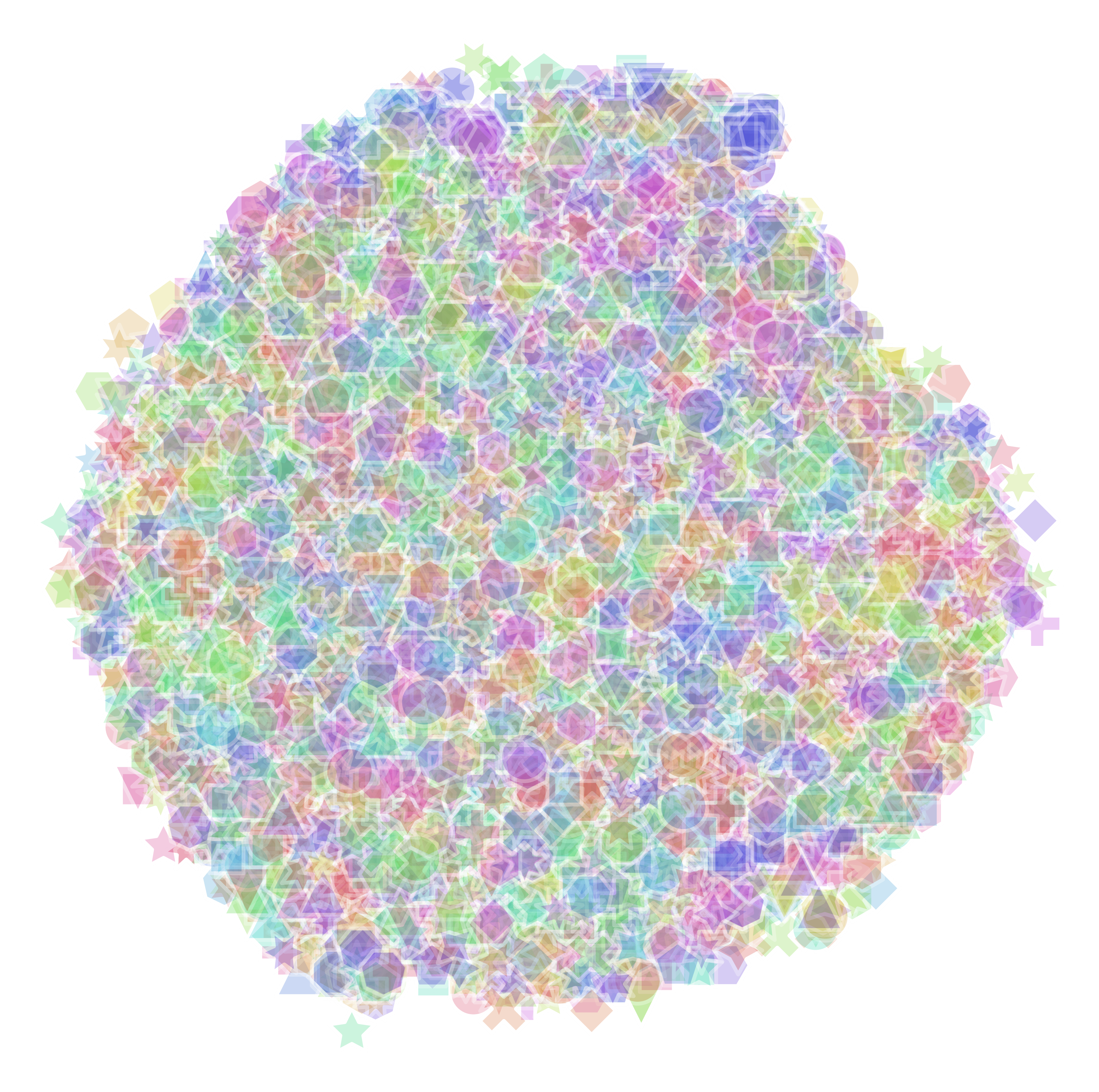}\label{fig:fedmp_before}}
    \hfill
	\subfigure[+\mpm (${\bm z}^g_i$, A).]{\includegraphics[width=0.16\linewidth]{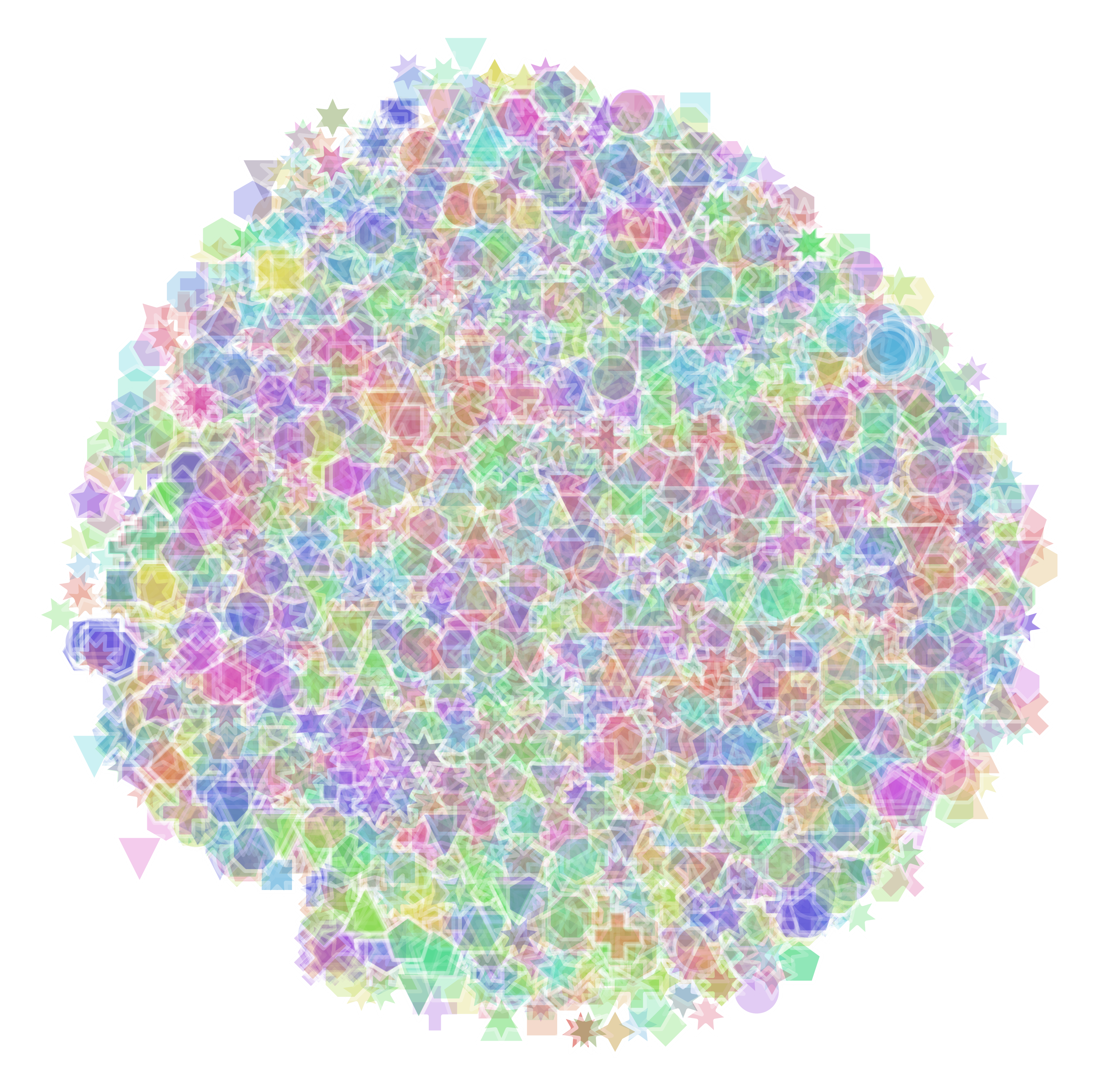}\label{fig:fedmp_after}}
	\subfigure[+\mpm (${\bm z}_i$, B).]{\includegraphics[width=0.16\linewidth]{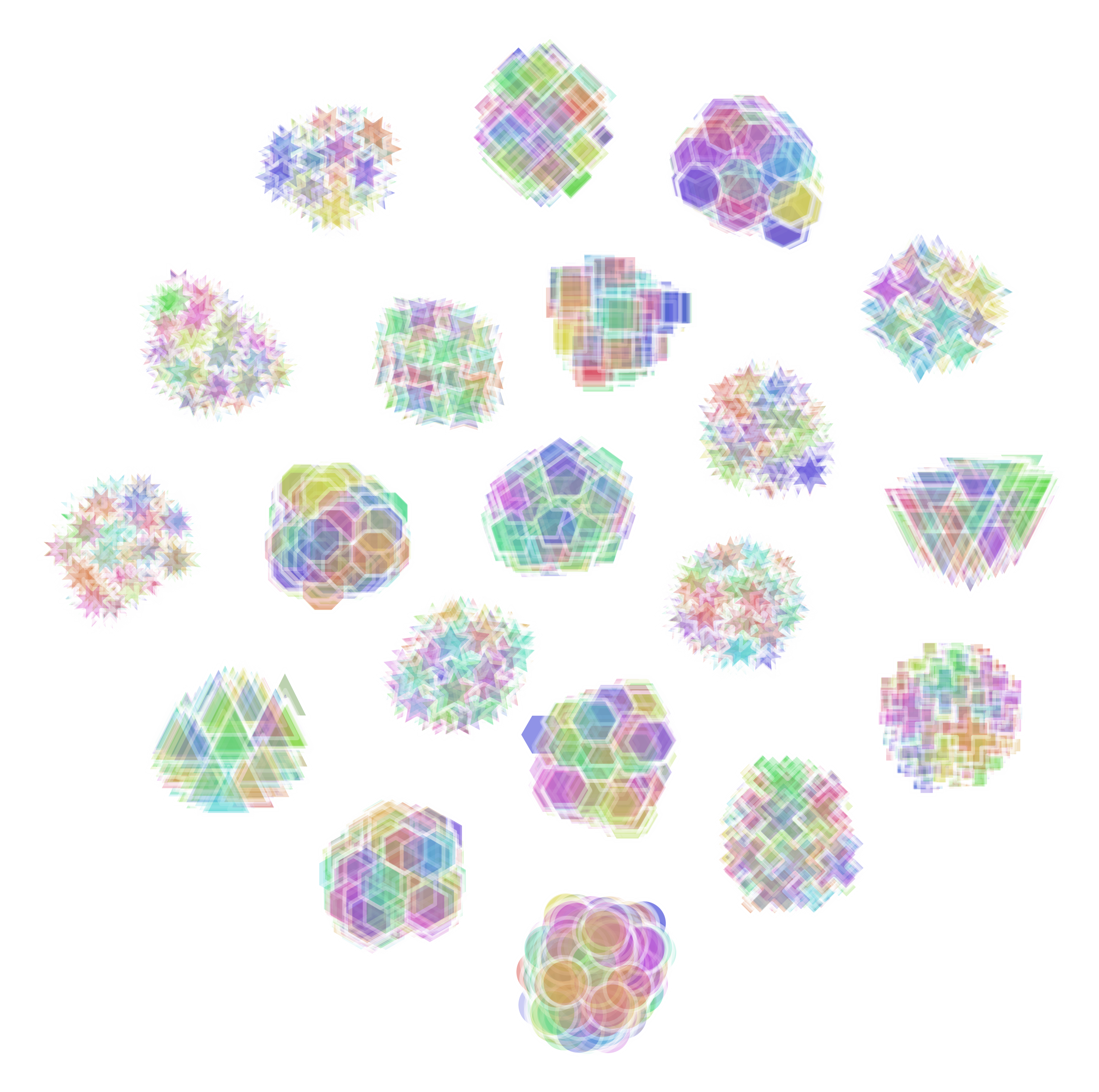}\label{fig:fedmpmp_before}}
    \hfill
	\subfigure[+\mpm (${\bm z}_i$, A).]{\includegraphics[width=0.16\linewidth]{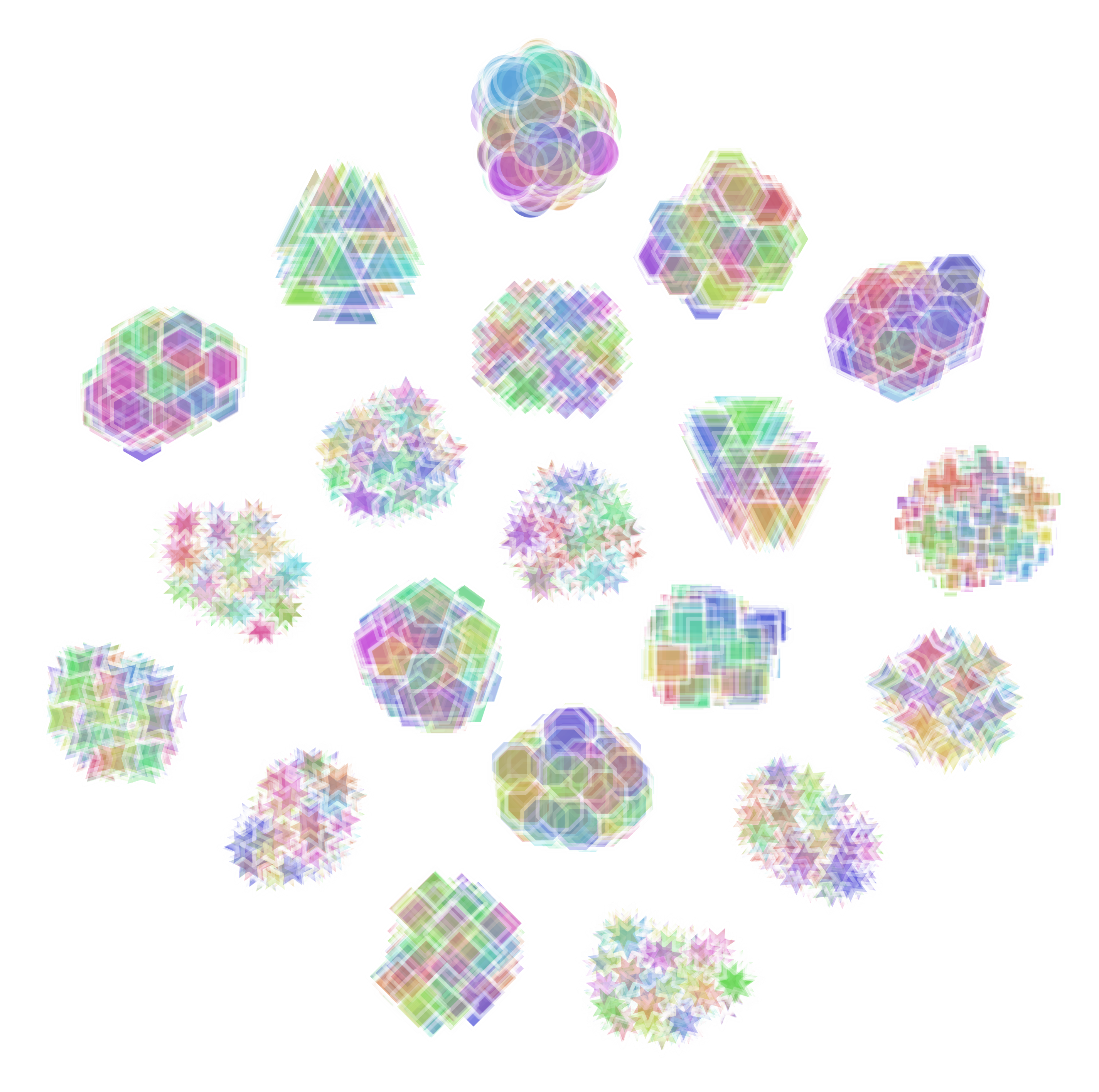}\label{fig:fedmpmp_after}}
	\caption{t-SNE visualization for representations on Tiny-ImageNet (200 labels). ``B'' and ``A'' denote ``before local training'' and ``after local training'', respectively. We use \textit{color} and \textit{shape} to distinguish \textit{labels} and \textit{clients}, respectively. \textit{Best viewed in color and zoom-in.} }
	\label{fig:rep}
\vspace{-20pt}
\end{figure}

We visualize the feature representations using t-SNE~\cite{van2008visualizing} in \Cref{fig:rep}. Compared to the representations outputted by the feature extractor in FedAvg, ${\bm z}^g_i$ in FedAvg+\mpm is no longer biased to the local data domain of each client after local training. With the personalized translation transformation \prbm, ${\bm z}_i$ can fit the local domain of each client either before or after local training. According to \Cref{fig:fed_after}, \Cref{fig:fedmpmp_before} and \Cref{fig:fedmpmp_after}, ${\bm z}_i$ in FedAvg+\mpm can fit the local domain better than FedAvg.

\subsubsection{Ablation Study for \mpm}

\begin{table}[h]
  \centering
  \caption{The MDL (bits, $\downarrow$) and test accuracy (\%, $\uparrow$) when adding \mpm to FedAvg on Tiny-ImageNet using 4-layer CNN and ResNet-18 in the practical setting. }
  \resizebox{0.9\linewidth}{!}{
    \begin{tabular}{l|cccc|cccc}
    \toprule
    \textbf{Models} & \multicolumn{4}{c|}{\textbf{4-layer CNN}} & \multicolumn{4}{c}{\textbf{ResNet-18}} \\
    \midrule
    \textbf{Components} & FedAvg & +\mr & +\prbm & +\mpm & FedAvg & +\mr & +\prbm & +\mpm\\
    \midrule
    MDL & 3844 & 3643 & 3699 & \textbf{3625} & 3560 & 3460 & 3471 & \textbf{3454} \\
    Accuracy & 19.46 & 22.21 & 26.70 & \textbf{43.32} & 19.45 & 20.85 & 38.27 & \textbf{42.98} \\
    \bottomrule
    \end{tabular}}
  \label{tab:ablation}
\vspace{-5pt} \end{table}

We further study the contribution of \mr and \prbm in terms of generalization and personalization abilities by applying only one of them to FedAvg. From \Cref{tab:ablation}, we find that for 4-layer CNN and ResNet-18, +\mpm gives a larger improvement in both MDL and accuracy than just using \mr or \prbm, which suggests that \mr and \prbm can boost each other in bi-directional knowledge transfer. The contribution of \mr is greater than that of \prbm in improving the generalization ability in MDL, while +\prbm gains more accuracy improvement for personalization ability than \mr. 
% +\prbm gets more improvement in ResNet-18 since batch normalization (BN)~\cite{ioffe2015batch} plays a role similar to \mr through normalization and recovering. However, BN also learns biased information during local training like other DNN layers, so \mpm can also improve ResNet-18 obviously. 

% Since batch normalization (BN)~\cite{ioffe2015batch} can normalize the mean and variance of input data on all clients through server aggregation, it plays a role similar to \mr, the improvement of +\mr in 4-layer CNN is greater than ResNet-18 and +\prbm get more improvement in ResNet-18. However, \mr can still improve ResNet-18 because the trainable BN also learns biased information during local training like other DNN layers. 
% With BN, ResNet-18 has better generalization ability than 4-layer CNN, thus achieving lower MDL but leaving less room for \prbm to improve personalization ability. 

\subsubsection{Privacy-Preserving Ability}
\label{sec:privacy}

\begin{table}[h]
  \centering
  \caption{The test accuracy (\%, $\uparrow$) using FedAvg+\mpm on TINY in the practical setting with noise.}
  \resizebox{!}{!}{
    \begin{tabular}{c|cccc|cccc}
    \toprule
     & \multicolumn{4}{c|}{$q=0.2$} & \multicolumn{4}{c}{\textbf{$s=0.05$}}\\
    \midrule
    Original & $s=0.05$ & $s=0.5$ & $s=1$ & $s=5$ & $q=0.1$ & $q=0.5$ & $q=0.8$ & $q=0.9$ \\
    \midrule    
    43.32 & 44.10 & 44.15 & 43.78 & 36.27 & 43.81 & \textbf{44.45} & 43.30 & 41.75 \\
    \bottomrule
    \end{tabular}}
  \label{tab:privacy}
\vspace{-5pt} \end{table}

Following FedPAC~\cite{tan2022federated}, we add Gaussian noise to client-specific means $\bar{{z}}^g_1, \ldots, \bar{{z}}^g_N$ with a scale parameter ($s$) for the noise distribution and perturbation coefficient ($q$) for the noise. Adding the unbiased noise sampled from one distribution is beneficial for representation bias elimination and can further improve the performance of \mpm to some extent, as shown in \Cref{tab:privacy}. Besides, adding too much noise can also bring an accuracy decrease. However, setting $s=0.05$ and $q=0.2$ is sufficient to ensure privacy protection according to FedPCL.

\subsubsection{\mpm Improves Other Traditional Federated Learning Methods}

\begin{table}[h]
  \centering
  \caption{The MDL (bits, $\downarrow$) and test accuracy (\%, $\uparrow$) before and after adding \mpm to traditional FL methods on Cifar100, Tiny-ImageNet, and AG News in the practical setting. TINY and TINY* represent using 4-layer CNN and ResNet-18 on Tiny-ImageNet, respectively. }
  \resizebox{\linewidth}{!}{
    \begin{tabular}{l|cccc|cccc}
    \toprule
    \textbf{Metrics} & \multicolumn{4}{c|}{\textbf{MDL}} & \multicolumn{4}{c}{\textbf{Accuracy}}\\
    \midrule
    \textbf{Datasets} & Cifar100 & TINY & TINY* & AG News & Cifar100 & TINY & TINY* & AG News\\
    \midrule
    SCAFFOLD~\cite{karimireddy2020scaffold} & 1499 & 3661 & 3394 & 1931 & 33.08 & 23.26 & 24.90 & 88.13 \\
    FedProx~\cite{MLSYS2020_38af8613} & 1523 & 3701 & 3570 & 2092 & 31.99 & 19.37 & 19.27 & 87.21 \\
    MOON~\cite{li2021model} & 1516 & 3696 & 3536 & 1836 & 32.37 & 19.68 & 19.02 & 84.14 \\
    FedGen~\cite{zhu2021data} & 1506 & 3675 & 3551 & 1414 & 30.96 & 19.39 & 18.53 & 89.86 \\
    \midrule
    SCAFFOLD+\mpm & \textbf{1434} & \textbf{3549} & \textbf{3370} & \textbf{1743} & \textbf{63.61} & \textbf{45.55} & \textbf{45.09} & \textbf{96.73} \\
    FedProx+\mpm & \textbf{1439} & \textbf{3587} & \textbf{3490} & \textbf{1689} & \textbf{63.22} & \textbf{42.28} & \textbf{41.45} & \textbf{96.62} \\
    MOON+\mpm & \textbf{1432} & \textbf{3580} & \textbf{3461} & \textbf{1683} & \textbf{63.26} & \textbf{43.43} & \textbf{41.10} & \textbf{96.68} \\
    FedGen+\mpm & \textbf{1426} & \textbf{3563} & \textbf{3488} & \textbf{1098} & \textbf{63.26} & \textbf{42.54} & \textbf{41.87} & \textbf{97.16} \\
    \bottomrule
    \end{tabular}}
  \label{tab:tFL}
\vspace{-5pt} \end{table}

A large number of FL methods design algorithms based on the famous FedAvg~\cite{mcmahan2017communication, kairouz2019advances, tan2022towards}. Although we describe \mpm based on FedAvg for example, \mpm can also be applied to other traditional FL methods to improve their generalization and personalization abilities. Here, we apply \mpm to another four representative traditional FL methods: SCAFFOLD~\cite{karimireddy2020scaffold}, FedProx~\cite{MLSYS2020_38af8613}, MOON~\cite{li2021model}, and FedGen~\cite{zhu2021data}. They belong to four categories: update-correction-based FL, regularization-based FL, model-split-based FL, and knowledge-distillation-based FL, respectively. In \Cref{tab:tFL}, \mpm promotes traditional FL methods by at most \textbf{\textcolor{green_}{-22.35\%}} in MDL (bits) and \textbf{\textcolor{green_}{+32.30}} in accuracy (\%), respectively. Based on the results of \Cref{tab:ablation} and \Cref{tab:tFL} on Tiny-ImageNet, FedAvg+\mpm achieves lower MDL and higher accuracy than close methods MOON and FedGen. 

\subsection{Comparison with Personalized Federated Learning Methods}

% To further show the superiority of the \mpm-equipped traditional FL methods to existing pFL methods, we compare the representative FedAvg+\mpm with ten pFL methods that are proposed based on FedAvg including Per-FedAvg~\cite{NEURIPS2020_24389bfe}, pFedMe~\cite{t2020personalized}, Ditto~\cite{li2021ditto}, FedPer~\cite{arivazhagan2019federated}, FedRep~\cite{collins2021exploiting}, FedRoD~\cite{chen2021bridging}, FedBABU~\cite{ohfedbabu}, APFL~\cite{deng2020adaptive}, FedFomo~\cite{zhang2020personalized}, and APPLE~\cite{ijcai2022p301}. For FedBABU, we show the test accuracy after post-FL fine-tuning for additional 10 local epochs here. 
% In the following, we show the results in personalization ability, communication and computation overhead, privacy-preserving ability, and Generalization Ability on New Clients. 

\subsubsection{Personalization Ability on Various Datasets}

\begin{table}[h] \vspace{-5pt}
  \centering
  \caption{The test accuracy (\%, $\uparrow$) of pFL methods in two statistically heterogeneous settings. Cifar100$^\dag$ represents the experiment with 100 clients and joining ratio $\rho=0.5$ on Cifar100. }
  \resizebox{\linewidth}{!}{
    \begin{tabular}{l|ccc|cccccc}
    \toprule
    \textbf{Settings} & \multicolumn{3}{c|}{\textbf{Pathological setting}} & \multicolumn{6}{c}{\textbf{Practical setting}} \\
    \midrule
     & FMNIST & Cifar100 & TINY & FMNIST & Cifar100 & Cifar100$^\dag$ & TINY & TINY* & AG News\\
    \midrule
    Per-FedAvg~\cite{NEURIPS2020_24389bfe} & 99.18 & 56.80 & 28.06 & 95.10 & 44.28 & 38.28 & 25.07 & 21.81 & 87.08 \\
    pFedMe~\cite{t2020personalized} & 99.35 & 58.20 & 27.71 & 97.25 & 47.34 & 31.13 & 26.93 & 33.44 & 87.08 \\
    Ditto~\cite{li2021ditto} & 99.44 & 67.23 & 39.90 & 97.47 & 52.87 & 39.01 & 32.15 & 35.92 & 91.89 \\
    FedPer~\cite{arivazhagan2019federated} & 99.47 & 63.53 & 39.80 & 97.44 & 49.63 & 41.21 & 33.84 & 38.45 & 91.85 \\
    FedRep~\cite{collins2021exploiting} & 99.56 & 67.56 & 40.85 & 97.56 & 52.39 & 41.51 & 37.27 & 39.95 & 92.25 \\
    FedRoD~\cite{chen2021bridging} & 99.52 & 62.30 & 37.95 & 97.52 & 50.94 & 48.56 & 36.43 & 37.99 & 92.16 \\
    FedBABU~\cite{ohfedbabu} & 99.41 & 66.85 & 40.72 & 97.46 & 55.02 & 52.07 & 36.82 & 34.50 & 95.86 \\
    APFL~\cite{deng2020adaptive} & 99.41 & 64.26 & 36.47 & 97.25 & 46.74 & 39.47 & 34.86 & 35.81 & 89.37 \\
    FedFomo~\cite{zhang2020personalized} & 99.46 & 62.49 & 36.55 & 97.21 & 45.39 & 37.59 & 26.33 & 26.84 & 91.20 \\
    APPLE~\cite{ijcai2022p301} & 99.30 & 65.80 & 36.22 & 97.06 & 53.22 & --- & 35.04 & 39.93 & 84.10 \\
    \midrule
    FedAvg & 80.41 & 25.98 & 14.20 & 85.85 & 31.89 & 28.81 & 19.46 & 19.45 & 87.12 \\
    FedAvg+\mpm & \textbf{99.74} & \textbf{73.38} & \textbf{42.89} & \textbf{97.69} & \textbf{64.39} & \textbf{63.43} & \textbf{43.32} & \textbf{42.98} & \textbf{96.87} \\
    \bottomrule
    \end{tabular}}
  \label{tab:per}
\vspace{-5pt} \end{table}

To further show the superiority of the \mpm-equipped traditional FL methods to existing pFL methods, we compare the representative FedAvg+\mpm with ten SOTA pFL methods, as shown in \Cref{tab:per}. Note that APPLE is designed for cross-silo scenarios and assumes $\rho=1$. For Per-FedAvg and FedBABU, we show the test accuracy after post-FL fine-tuning. FedAvg+\mpm improves FedAvg at most \textbf{\textcolor{green_}{+47.40}} on Cifar100 in the pathological setting and outperforms the best SOTA pFL methods by up to \textbf{\textcolor{green_}{+11.36}} on Cifar100$^\dag$ including the fine-tuning-based methods that require additional post-FL effort. 

\subsubsection{Personalization Ability Under Various Heterogeneous Degrees}

Following prior methods~\cite{li2021model, NEURIPS2020_18df51b9}, we also evaluate FedAvg+\mpm with different $\beta$ on Tiny-ImageNet using 4-layer CNN to study the influence of heterogeneity, as shown in \Cref{tab:misc}. Most pFL methods are specifically designed for extremely heterogeneous scenarios and can achieve high accuracy at $\beta=0.01$, but some of them cannot maintain the advantage compared to FedAvg in moderate scenarios. However, FedAvg+\mpm can automatically adapt to all these scenarios without tuning. 

\begin{table}[h]
  \centering
  \caption{The test accuracy (\%, $\uparrow$) and computation overhead ($\downarrow$) of pFL methods. }
  \resizebox{\linewidth}{!}{
    \begin{tabular}{l|ccc|cc|rr}
    \toprule
    \textbf{Items} & \multicolumn{3}{c|}{\textbf{Heterogeneity}} & \multicolumn{2}{c|}{\textbf{pFL+\mr}} & \multicolumn{2}{c}{\textbf{Overhead}}\\
    \midrule
     & $\beta=0.01$ & $\beta=0.5$ & $\beta=5$ & Accuracy & Improvement & Total time & Time/iteration\\
    \midrule
    Per-FedAvg~\cite{NEURIPS2020_24389bfe} & 39.39 & 16.36 & 12.08 & --- & --- & 121 min & 3.56 min\\
    pFedMe~\cite{t2020personalized} & 41.45 & 17.48 & 4.03 & --- & --- & 1157 min & 10.24 min\\
    Ditto~\cite{li2021ditto} & 50.62 & 18.98 & 21.79 & 42.82 & 10.67 & 318 min & 11.78 min\\
    FedPer~\cite{arivazhagan2019federated} & 51.83 & 17.31 & 9.61 & 41.78 & 7.94 & 83 min & 1.92 min\\
    FedRep~\cite{collins2021exploiting} & 55.43 & 16.74 & 8.04 & 41.28 & 4.01 & 471 min & 4.09 min\\
    FedRoD~\cite{chen2021bridging} & 49.17 & 23.23 & 16.71 & 42.74 & 6.31 & 87 min & 1.74 min\\
    FedBABU~\cite{ohfedbabu} & 53.97 & 23.08 & 15.42 & 38.17 & 1.35 & 811 min & 1.58 min\\
    APFL~\cite{deng2020adaptive} & 49.96 & 23.31 & 16.12 & 39.22 & 4.36 & 156 min & 2.74 min\\
    FedFomo~\cite{zhang2020personalized} & 46.36 & 11.59 & 14.86 & 29.51 & 3.18 & 193 min & 2.72 min\\
    APPLE~\cite{ijcai2022p301} & 48.04 & 24.28 & 17.79 & --- & --- & 132 min & 2.93 min\\
    \midrule
    FedAvg & 15.70 & 21.14 & 21.71 & --- & --- & 365 min & 1.59 min\\
    FedAvg+\mpm & \textbf{57.52} & \textbf{32.61} & \textbf{25.55} & --- & --- & 171 min & 1.60 min\\
    \bottomrule
    \end{tabular}}
  \label{tab:misc}
\vspace{-5pt} \end{table}

\subsubsection{\mr Improves Personalized Federated Learning Methods}

Since pFL methods already create personalized models or modules in their specific ways, applying personalized \prbm to the local model might be against their philosophy. To prevent this, we only apply the \mr to pFL methods. Besides, the local training schemes (\eg, meta-learning) in Per-FedAvg, pFedMe, and APPLE are different from the simple SGD in FedAvg, which requires modification of the mean calculation in \mr, so we do not apply \mr to them. 
% the local model training for the pFL methods that follow the local training scheme of FedAvg, and report results in \Cref{tab:misc}. 
According to \Cref{corollary:mr}, \mr can promote the local-to-global knowledge transfer between server and client. Therefore, pFL methods can benefit more from a better global model achieving higher accuracy on Tiny-ImageNet with the 4-layer CNN, as shown in \Cref{tab:misc}. However, their \mr-equipped variants perform worse than FedAvg+\mpm (\Cref{tab:per}, TINY) since the representation bias still exists without using \prbm. 

\subsubsection{Computation Overhead}

We evaluate FedAvg+\mpm in total time and time per iteration on Tiny-ImageNet using ResNet-18, as shown in \Cref{tab:misc}. The evaluation task for one method monopolizes one identical machine. FedAvg, FedBABU, and FedAvg+\mpm cost almost the same and have the lowest time per iteration among these methods, but FedAvg+\mpm requires less total time than FedAvg and FedBABU. Note that the fine-tuning time for FedBABU is not included in \Cref{tab:misc}. Since pFedMe and Ditto train an additional personalized model on each client, they cost plenty of time per iteration. 

\section{Conclusion}

Due to the naturally existing statistical heterogeneity and the biased local data domains on each client, FL suffers from representation bias and representation degeneration problems. To improve the generalization and personalization abilities for FL, we propose a general framework \mpm including two modules \prbm and \mr, with a theoretical guarantee. Our \mpm can promote the bi-directional knowledge transfer in each iteration, thus improving both generalization and personalization abilities. Besides, we conduct extensive experiments to show the general applicability of \mpm to existing FL methods and the superiority of the representative FedAvg+\mpm to ten SOTA pFL methods in various scenarios. 
% It is valuable to view FL from a knowledge transfer perspective. 

\begin{ack}
This work was partially supported by the Program of Technology Innovation of the Science and Technology Commission of Shanghai Municipality (Granted No. 21511104700 and 22DZ1100103). This work was also supported in part by the Shanghai Key Laboratory of Scalable Computing and Systems, National Key R\&D Program of China (2022YFB4402102), Internet of Things special subject program, China Institute of IoT (Wuxi), Wuxi IoT Innovation Promotion Center (2022SP-T13-C), Industry-university-research Cooperation Funding Project from the Eighth Research Institute in China Aerospace Science and Technology Corporation (Shanghai) (USCAST2022-17), Intel Corporation (UFunding 12679), and the cooperation project from Ant Group (``Wasm-enabled Managed language in security restricted scenarios''). 
\end{ack}

%%%%%%%%%%%%%%%%%%%%%%%%%%%%%%%%
% \clearpage
\small
\bibliographystyle{plainnat}
\bibliography{main}

\clearpage
\appendix

\setcounter{theorem}{0}
\setcounter{corollary}{0}
\setcounter{definition}{0}

We provide more details and results about our work in the appendices. Here are the contents:

\begin{itemize}
    % \item \Cref{sec:algo}: The algorithm of the representative FedAvg+\mpm. 
    \item \Cref{sec:related}: The extended version of the Related Work section in the main body. 
    \item \Cref{sec:theo}: Proofs of Corollary 1 and Corollary 2. 
    \item \Cref{sec:setting}: More details about experimental settings. 
    \item \Cref{sec:addexp}: Additional experiments (\eg, a real-world application).
    \item \Cref{sec:impact}: Broader impacts of our proposed method.
    \item \Cref{sec:limit}: Limitations of our proposed method. 
    \item \Cref{sec:disvis}: Data distribution visualizations for different scenarios in our experiments. 
\end{itemize}

\section{Related Work}
\label{sec:related}

As the number of users and sensors rapidly increases with massive growing services on the Internet, the privacy concerns about private data also draw increasing attention of researchers~\cite{talesh2018data, kairouz2019advances, tesfay2018read}. Then a new distributed machine learning paradigm, federated learning (FL), comes along with the privacy-preserving and collaborative learning abilities~\cite{yang2019federated, kairouz2019advances, mcmahan2017communication}. Although there are horizontal FL~\cite{mcmahan2017communication, MLSYS2020_38af8613, yang2019federated}, vertical FL~\cite{luo2021feature, yang2019federated, qi2022fairvfl}, federated transfer learning\cite{liu2020secure, chen2020fedhealth}, \etc, we focus on the popular horizontal FL and call it FL for short in this paper. 

Traditional FL methods concentrate on learning a single global model among a  server and clients, but it suffers an accuracy decrease under statistically heterogeneous scenarios, which are common scenarios in practice~\cite{mcmahan2017communication, t2020personalized, li2021ditto, zhang2022fedala}. Then, many FL methods propose learning personalized models (or modules) for each client besides learning the global model. These FL methods are specifically called personalized FL (pFL) methods~\cite{tan2022towards, collins2021exploiting, NEURIPS2020_24389bfe}. 

\subsection{Traditional Federated Learning}

FL methods perform machine learning through iterative communication and computation on the server and clients. To begin with, we describe the FL procedure in one iteration based on FedAvg~\cite{mcmahan2017communication}, which is a famous FL method and a basic framework for later FL methods. The FL procedure includes five stages: (1) A server selects a group of clients to join FL in this iteration and sends the current global model to them; (2) these clients receive the global model and initialize their local model by overwriting their local model with the parameters in the global model; (3) these clients train their local models on their own private local data, respectively; (4) these clients send the trained local models to the server; (5) the server receives client models and aggregates them through weighted averaging on model parameters to obtain a new global model. 

Then, massive traditional FL methods are proposed in the literature to improve FedAvg regarding privacy-preserving~\cite{li2020preserving, zhu2020federated, mothukuri2021survey}, accuracy~\cite{karimireddy2020scaffold, zhu2021data, li2021model}, fairness~\cite{yu2020fairness, huang2020efficiency}, overhead~\cite{luping2019cmfl, konevcny2016federated, hao2019efficient}, \etc. Here, we focus on the representative traditional FL methods that handle the heterogeneity issues in four categories: update-correction-based FL~\cite{karimireddy2020scaffold, niu2022federated, gao2022feddc}, regularization-based FL~\cite{MLSYS2020_38af8613, acarfederated, kim2022multi, cheng2022differentially}, model-split-based FL~\cite{li2021model, jiang2022harmofl}, and knowledge-distillation-based FL~\cite{zhu2021data, zhang2022fine, gong2022preserving, huang2022learn}.

Among \textbf{update-correction-based FL} methods, SCAFFOLD~\cite{karimireddy2020scaffold} witnesses the client-drift phenomenon of FedAvg under statistically heterogeneous scenarios due to local training and proposes correcting local update through control variates for each model parameter. Among \textbf{regularization-based FL} methods, FedProx~\cite{MLSYS2020_38af8613} modifies the local objective on each client by adding a regularization term to keep local model parameters close to the global model during local training in an element-wise manner. Among \textbf{model-split-based FL} methods, MOON~\cite{li2021model} observes that local training degenerates representation quality, so it adds a contrastive learning term to let the representations outputted by the local feature extractor be close to the ones outputted by the received global feature extractor given each input during local training. However, input-wise contrastive learning relies on biased local data domains, so MOON still suffers from representation bias. Among \textbf{knowledge-distillation-based FL} methods, FedGen~\cite{zhu2021data} learns a generator on the server to produce additional representations, shares the generator among clients, and locally trains the classifier with the combination of the representations outputted by the local feature extractor and the additionally generated representations. In this way, FedGen can reduce the heterogeneity among clients with the augmented representations from the shared generator via knowledge distillation. However, it only considers the local-to-global knowledge transfer for the single global model learning and additionally brings communication and computation overhead for learning and transmitting the generator. 

\subsection{Personalized Federated Learning}

Different from traditional FL, pFL additionally learns personalized models (or modules) besides the global model. In this paper, we consider pFL methods in four categories: meta-learning-based pFL~\cite{NEURIPS2020_24389bfe, chen2018federated}, regularization-based pFL~\cite{t2020personalized, li2021ditto}, personalized-aggregation-based pFL~\cite{deng2020adaptive, ijcai2022p301, zhang2020personalized}, and model-split-based pFL~\cite{arivazhagan2019federated, collins2021exploiting, chen2021bridging, ohfedbabu}. 

\noindent\textbf{Meta-learning-based pFL.\ \ } Meta-learning is a technique that trains deep neural networks (DNNs) on a given dataset for quickly adapting to other datasets with only a few steps of fine-tuning, \eg, MAML~\cite{finn2017model}. By integrating MAML into FL, Per-FedAvg~\cite{NEURIPS2020_24389bfe} updates the local models like MAML to capture the learning trends of each client and then aggregates the learning trends by averaging on the server. It obtains personalized models by fine-tuning the global model for each client. Similar to Per-FedAvg, FedMeta~\cite{chen2018federated} also introduces MAML on each client during training and fine-tuning the global model for evaluation. However, it is hard for these meta-learning-based pFL methods to find a consensus learning trend through averaging under statistically heterogeneous scenarios. 

\noindent\textbf{Regularization-based pFL.\ \ } Like FedProx, pFedMe~\cite{t2020personalized} and Ditto~\cite{li2021ditto} also utilize the regularization technique, but they modify the objective for additional personalized model training rather than the one for local model training. In  pFedMe and Ditto, each client owns two models: the local model that is trained for global model aggregation and the personalized model that is trained for personalization. Specifically, pFedMe regularizes the model parameters between the personalized model and the local model during training while Ditto regularizes the model parameters between the personalized model and the received global model. Besides, Ditto simply trains the local model similar to FedAvg while pFedMe trains the local model based on the personalized model. Although the local model is initialized by the global model, but the initialized local model gradually loses global information during local training. Thus, the personalized model in Ditto can be aware of more global information than the one in pFedMe. Both pFedMe and Ditto require additional memory space to store the personalized model and double the computation resources at least to train both the local model and the personalized model. 

\noindent\textbf{Personalized-aggregation-based pFL.\ \ } These pFL methods adaptively aggregate the global model and local model according to the local data on each client, \eg, APFL~\cite{deng2020adaptive}, or directly generate the personalized model using other client models through personalized aggregation on each client, \eg, FedFomo~\cite{zhang2020personalized} and APPLE~\cite{ijcai2022p301}. Specifically, APFL aggregates the parameters in the global model and the local model with weighted averaging and adaptively updates the scalar weight based on the gradients. On each client, FedFomo generates the client-specific aggregating weights for the received client models through first-order approximation while APPLE adaptively learns these weights based on the local data. Both FedFomo and APPLE require multiple communication overhead than other FL methods, but FedFomo costs less computation overhead than APPLE attributed to approximation. 

\noindent\textbf{Model-split-based pFL.\ \ } These pFL methods split a given model into a feature extractor and a classifier. They treat the feature extractor and the classifier differently. 
Concretely, FedPer~\cite{arivazhagan2019federated} and FedRep~\cite{collins2021exploiting} keep the classifier locally on each client. FedPer trains the feature extractor and the classifier together while FedRep first fine-tunes the classifier and then trains the feature extractor in each iteration. For FedPer and FedRep, the feature extractor intends to extract representations to cater to these personalized classifiers, thus reducing the generic representation quality. FedRoD~\cite{chen2021bridging} trains the local model with the balanced softmax (BSM) loss function~\cite{ren2020balanced} and simultaneously learns an additional personalized classifier for each client. However, the BSM loss is useless for missing labels on each client while label missing is a common situation in statistically heterogeneous scenarios~\cite{NEURIPS2020_18df51b9, zhang2022fedala, zhang2020personalized}. Moreover, the uniform label distribution modified by the BSM cannot reflect the original distribution. The above pFL methods learn personalized models (or modules) in FL, but FedBABU~\cite{ohfedbabu} firstly trains the global feature extractor with the frozen classifier during the FL process, then it fine-tunes the global model on each client after FL to obtain personalized models. However, this post-FL fine-tuning is beyond the scope of FL. Almost all the FL methods have multiple fine-tuning variants, \eg, fine-tuning the whole model or only a part of the model. Furthermore, training the feature extractor with the naive and randomly initialized classifier in FL has an uncontrollable risk due to randomness.

\section{Theoretical Derivations}
\label{sec:theo}

\subsection{Notations and Preliminaries}

Following prior arts~\cite{deng2020adaptive, zhu2021data, tan2022towards, mansour2020three}, we consider a binary classification problem in FL here. Recall that $\mathcal{X} \subset \mathbb{R}^D$ is an input space, $\mathcal{Z} \subset \mathbb{R}^K$ is a representation space, and $\mathcal{Y} \subset \{0, 1\}$ is a label space. Let $\mathcal{F}: \mathcal{X} \mapsto \mathcal{Z}$ be a representation function that maps from the input space to the representation space. We denote $\mathcal{D} := \langle \mathcal{U}, c^* \rangle$ as a data domain where the distribution $\mathcal{U} \subseteq \mathcal{X}$ and $c^*: \mathcal{X} \mapsto \mathcal{Y}$ is a ground-truth labeling function. $\tilde{\mathcal{U}}$ is the induced distribution of $\mathcal{U}$ over the representation space $\mathcal{Z}$ under $\mathcal{F}$~\cite{ben2006analysis}, \ie, $\tilde{\mathcal{U}} \subseteq \mathcal{Z}$, that satisfies
\begin{equation}
    \mathbb{E}_{{\bm z} \sim \tilde{\mathcal{U}}}\left[\mathcal{B}\left({\bm z}\right)\right] = \mathbb{E}_{{\bm x} \sim \mathcal{U}}\left[\mathcal{B}\left(\mathcal{F}\left({\bm x}\right)\right)\right], 
\end{equation}
where $\mathcal{B}$ is a probability event. 
Given fixed but unknown $\mathcal{U}$ and $c^*$, the learning task on one domain is to choose a representation function $\mathcal{F}$ and a hypothesis class $\mathcal{H}\subseteq \{h: \mathcal{Z} \mapsto \mathcal{Y}\}$ to approximate the function $c^*$. 

Then, we provide the definition and theorem from \citet{ben2010theory, kifer2004detecting, ben2006analysis, blitzer2007learning} under their assumptions:

\begin{definition}
    If a space $\mathcal{Z}$ with $\tilde{\mathcal{U}}^a$ and $\tilde{\mathcal{U}}^b$ distributions over $\mathcal{Z}$, let $\mathcal{H}$ be a hypothesis class on $\mathcal{Z}$ and $\mathcal{Z}_h \subseteq \mathcal{Z}$ be the subset with characteristic function $h$, the $\mathcal{H}$-divergence between $\tilde{\mathcal{U}}^a$ and $\tilde{\mathcal{U}}^b$ is 
    $$
    d_{\mathcal{H}}\left(\tilde{\mathcal{U}}^a, \tilde{\mathcal{U}}^b\right) = 2\sup_{h \in \mathcal{H}} \left| {\rm Pr}_{\tilde{\mathcal{U}}^a}\left[\mathcal{Z}_h\right] - {\rm Pr}_{\tilde{\mathcal{U}}^b}\left[\mathcal{Z}_h\right] \right|,
    $$
    where $\mathcal{Z}_h = \{{\bm z}\in \mathcal{Z}: h\left({\bm z}\right)=1\}, h \in \mathcal{H}$.\label{def}
\end{definition}

\cref{def} implies that $d_{\mathcal{H}}\left(\tilde{\mathcal{U}}^a, \tilde{\mathcal{U}}^b\right) = d_{\mathcal{H}}\left(\tilde{\mathcal{U}}^b, \tilde{\mathcal{U}}^a\right)$.  

\begin{theorem}
    Consider a source domain $\mathcal{D}_S$ and a target domain $\mathcal{D}_T$. Let $\mathcal{D}_S = \langle \mathcal{U}_S, c^* \rangle$ and $\mathcal{D}_T = \langle \mathcal{U}_T, c^* \rangle$, where $\mathcal{U}_S \subseteq \mathcal{X}$, $\mathcal{U}_T \subseteq \mathcal{X}$, and $c^*: \mathcal{X} \mapsto \mathcal{Y}$ is a ground-truth labeling function. Let $\mathcal{H}$ be a hypothesis space of VC dimension $d$ and $h: \mathcal{Z} \mapsto \mathcal{Y}, \forall \ h \in \mathcal{H}$. Given a feature extraction function $\mathcal{F}: \mathcal{X} \mapsto \mathcal{Z}$ that shared between $\mathcal{D}_S$ and $\mathcal{D}_T$, a random labeled sample of size $m$ generated by applying $\mathcal{F}$ to a random sample from $\mathcal{U}_S$ labeled according to $c^*$, then for every $h\in \mathcal{H}$, with probability at least $1-\delta$: 
    $$ 
    \mathcal{L}_{\mathcal{D}_T}\left(h\right) \le \mathcal{L}_{\hat{\mathcal{D}}_S}\left(h\right) + \sqrt{\frac{4}{m} \left(d\log\frac{2em}{d} + \log\frac{4}{\delta}\right)} + d_{\mathcal{H}}\left(\tilde{\mathcal{U}}_S, \tilde{\mathcal{U}}_T\right) + \lambda, 
    $$
    where $\mathcal{L}_{\hat{\mathcal{D}}_S}$ is the empirical loss on $\mathcal{D}_S$, $e$ is the base of the natural logarithm, and $d_{\mathcal{H}}\left(\cdot, \cdot\right)$ is the $\mathcal{H}$-divergence between two distributions. $\tilde{\mathcal{U}}_S$ and $\tilde{\mathcal{U}}_T$ are the induced distributions of $\mathcal{U}_S$ and $\mathcal{U}_T$ under $\mathcal{F}$, respectively, s.t. $\mathbb{E}_{{\bm z} \sim \tilde{\mathcal{U}}_S} \left[\mathcal{B}\left({\bm z}\right)\right] = \mathbb{E}_{{\bm x} \sim \mathcal{U}_S} \left[\mathcal{B}\left(\mathcal{F}\left({\bm x}\right)\right)\right]$ given a probability event $\mathcal{B}$, and so for $\tilde{\mathcal{U}}_T$. $\tilde{\mathcal{U}}_S \subseteq \mathcal{Z}$ and $\tilde{\mathcal{U}}_T \subseteq \mathcal{Z}$. $\lambda := \min_{h} \mathcal{L}_{\mathcal{D}_S}\left(h\right) + \mathcal{L}_{\mathcal{D}_T}\left(h\right)$ denotes an oracle performance. \label{the:1}
\end{theorem}

The traditional FL methods, which focus on enhancing the performance of a global model, regard local domains $\mathcal{D}_i, i\in [N]$ and the virtual global domain $\mathcal{D}$ as the source domain and the target domain, respectively~\cite{zhu2021data}, which is called local-to-global knowledge transfer in this paper. In contrast, pFL methods that focus on improving the performance of personalized models regard $\mathcal{D}$ and $\mathcal{D}_i, i\in [N]$ as the source domain and the target domain, respectively~\cite{deng2020adaptive, tan2022towards, mansour2020three}. We call this kind of adaptation global-to-local knowledge transfer. 
The local-to-global knowledge transfer happens on the server while the global-to-local one occurs on the client. 

\subsection{Derivations of Corollary 1}

As we focus on the local-to-global knowledge transfer on the \textit{server side}, in the FL scenario, we can rewrite \cref{the:1} to 
\begin{theorem}
    Consider a local data domain $\mathcal{D}_i$ and a virtual global data domain $\mathcal{D}$. Let $\mathcal{D}_i = \langle \mathcal{U}_i, c^* \rangle$ and $\mathcal{D} = \langle \mathcal{U}, c^* \rangle$, where $\mathcal{U}_i \subseteq \mathcal{X}$ and $\mathcal{U} \subseteq \mathcal{X}$. Given a feature extraction function $\mathcal{F}: \mathcal{X} \mapsto \mathcal{Z}$ that shared between $\mathcal{D}_i$ and $\mathcal{D}$, a random labeled sample of size $m$ generated by applying $\mathcal{F}$ to a random sample from $\mathcal{U}_i$ labeled according to $c^*$, then for every $h\in \mathcal{H}$, with probability at least $1-\delta$: 
    $$ 
    \mathcal{L}_{\mathcal{D}}\left(h\right) \le \mathcal{L}_{\hat{\mathcal{D}}_i}\left(h\right) + \sqrt{\frac{4}{m} \left(d\log\frac{2em}{d} + \log\frac{4}{\delta}\right)} + d_{\mathcal{H}}\left(\tilde{\mathcal{U}}_i, \tilde{\mathcal{U}}\right) + \lambda_i, 
    $$
    where $\tilde{\mathcal{U}}_i$ and $\tilde{\mathcal{U}}$ are the induced distributions of $\mathcal{U}_i$ and $\mathcal{U}$ under $\mathcal{F}$, respectively. $\tilde{\mathcal{U}}_i \subseteq \mathcal{Z}$ and $\tilde{\mathcal{U}} \subseteq \mathcal{Z}$. $\lambda_i := \min_{h} \mathcal{L}_{\mathcal{D}_i}\left(h\right) + \mathcal{L}_{\mathcal{D}}\left(h\right)$ denotes an oracle performance. \label{the:2}
\end{theorem}

\begin{corollary}
    Consider a local data domain $\mathcal{D}_i$ and a virtual global data domain $\mathcal{D}$ for client $i$ and the server, respectively. Let $\mathcal{D}_i = \langle \mathcal{U}_i, c^* \rangle$ and $\mathcal{D} = \langle \mathcal{U}, c^* \rangle$, where $c^*: \mathcal{X} \mapsto \mathcal{Y}$ is a ground-truth labeling function. Let $\mathcal{H}$ be a hypothesis space of VC dimension $d$ and $h: \mathcal{Z} \mapsto \mathcal{Y}, \forall \ h \in \mathcal{H}$. When using \mpm, given a feature extraction function $\mathcal{F}^g: \mathcal{X} \mapsto \mathcal{Z}$ that shared between $\mathcal{D}_i$ and $\mathcal{D}$, a random labeled sample of size $m$ generated by applying $\mathcal{F}^g$ to a random sample from $\mathcal{U}_i$ labeled according to $c^*$, then for every $h^g\in \mathcal{H}$, with probability at least $1-\delta$: 
    $$ 
    \mathcal{L}_{\mathcal{D}}\left(h^g\right) \le \mathcal{L}_{\hat{\mathcal{D}}_i}\left(h^g\right) + \sqrt{\frac{4}{m} \left(d\log\frac{2em}{d} + \log\frac{4}{\delta}\right)} + d_{\mathcal{H}}\left(\tilde{\mathcal{U}}^g_i, \tilde{\mathcal{U}}^g\right) + \lambda_i, 
    $$
    where$\mathcal{L}_{\hat{\mathcal{D}}_i}$ is the empirical loss on $\mathcal{D}_i$, $e$ is the base of the natural logarithm, and $d_{\mathcal{H}}\left(\cdot, \cdot\right)$ is the $\mathcal{H}$-divergence between two distributions. $\lambda_i := \min_{h^g} \mathcal{L}_{\mathcal{D}}\left(h^g\right) + \mathcal{L}_{\mathcal{D}_i}\left(h^g\right)$, $\tilde{\mathcal{U}}^g_i \subseteq \mathcal{Z}$, $\tilde{\mathcal{U}}^g \subseteq \mathcal{Z}$, and $d_{\mathcal{H}}\left(\tilde{\mathcal{U}}^g_i, \tilde{\mathcal{U}}^g\right) \le d_{\mathcal{H}}\left(\tilde{\mathcal{U}}_i, \tilde{\mathcal{U}}\right)$. $\tilde{\mathcal{U}}^g_i$ and $\tilde{\mathcal{U}}^g$ are the induced distributions of $\mathcal{U}_i$ and $\mathcal{U}$ under $\mathcal{F}^g$, respectively. $\tilde{\mathcal{U}}_i$ and $\tilde{\mathcal{U}}$ are the induced distributions of $\mathcal{U}_i$ and $\mathcal{U}$ under $\mathcal{F}$, respectively. $\mathcal{F}$ is the feature extraction function in the original FedAvg without \mpm. \label{corollary:mr1}
\end{corollary}

\begin{proof}
    Computing $d_{\mathcal{H}}\left(\cdot, \cdot\right)$ is identical to learning a classifier to achieve a minimum error of discriminating between points sampled from $\tilde{\mathcal{U}}$ and $\tilde{\mathcal{U}}'$, \ie, a binary domain classification problem~\cite{ben2006analysis, ben2010theory}. The more difficult the domain classification problem is, the smaller $d_{\mathcal{H}}\left(\cdot, \cdot\right)$ is. 
    % \begin{lemma}
    %     Let $\mathcal{H}$ be a symmetric hypothesis class \left(for every $h\in \mathcal{H}$, the inverse hypothesis $1-h$ is also in $\mathcal{H}$\right). If $\tilde{\mathcal{S}}'$ and $\tilde{\mathcal{S}}$ are sampled from $\tilde{\mathcal{U}}'$ and $\tilde{\mathcal{U}}$ with size $m$ respectively,
    %     $$
    %     d_{\mathcal{H}}\left(\tilde{\mathcal{S}}', \tilde{\mathcal{S}}\right) = 2 \left(1 - \min_{h\in \mathcal{H}} \left[\frac{1}{m} \sum_{{\bm z}:h\left({\bm z}\right)=0} I\left[{\bm z} \in \tilde{\mathcal{S}}'\right] + \frac{1}{m} \sum_{{\bm z}:h\left({\bm z}\right)=1} I\left[{\bm z} \in \tilde{\mathcal{S}}\right]\right]\right),
    %     $$
    %     where $d_{\mathcal{H}}\left(\tilde{\mathcal{S}}', \tilde{\mathcal{S}}\right)$ is the empirical version of $d_{\mathcal{H}}\left(\tilde{\mathcal{U}}, \tilde{\mathcal{U}}'\right)$ and $I\left[{\bm z} \in \tilde{\mathcal{S}}\right]$ is a binary indicator which equals to $1$ if ${\bm z} \in \tilde{\mathcal{S}}$.
    % \end{lemma}
    Unfortunately, computing the error of the optimal hyperplane classifier for arbitrary distributions is a well-known NP-hard problem~\cite{ben2006analysis, ben2003difficulty}. Thus, researchers approximate the error by learning a linear classifier for the binary domain classification~\cite{blitzer2007biographies, blitzer2007learning, ben2003difficulty}. Inspired by previous approaches~\cite{balakrishnama1998linear, mika1999fisher, kuhn2013discriminant}, we consider using Linear Discriminant Analysis (LDA) for the binary domain classification. The discrimination ability of LDA is measured by the Fisher discriminant ratio (F1)~\cite{ho2002complexity, wang2011feature, cano2013analysis}
    $$
    F1\left(\tilde{\mathcal{U}}^a, \tilde{\mathcal{U}}^b\right) = \max_k\left[\frac{\left({\bm \mu}^k_{\tilde{\mathcal{U}}^a} - {\bm \mu}^k_{\tilde{\mathcal{U}}^b}\right)^2}{\left({\bm \sigma}^k_{\tilde{\mathcal{U}}^a}\right)^2 + \left({\bm \sigma}^k_{\tilde{\mathcal{U}}^b}\right)^2}\right], 
    $$
    where ${\bm \mu}^k_{\tilde{\mathcal{U}}^a}$ and $\left({\bm \sigma}^k_{\tilde{\mathcal{U}}^a}\right)^2$ are the mean and variance of the values in the $k$th dimension over $\tilde{\mathcal{U}}^a$. The smaller the Fisher discriminant ratio is, the less discriminative the two domains are. \cref{the:2} holds with every $h\in \mathcal{H}$, so we omit \prbm here. 
    $\mr\left(\bar{{\bm z}}^g_i, \bar{{\bm z}}^g\right)$ forces the local domain to be close to the global domain in terms of the mean value at each feature dimension in the feature representation independently, therefore, $\forall \ k \in \left[K\right]$, $${\bm \mu}^k_{\tilde{\mathcal{U}}^g_i} - {\bm \mu}^k_{\tilde{\mathcal{U}}^g} \le {\bm \mu}^k_{\tilde{\mathcal{U}}_i} - {\bm \mu}^k_{\tilde{\mathcal{U}}}.$$  As the feature extractors share the same structure with identical parameter initialization and the feature representations are extracted from the same data domain $\mathcal{D}_i$ ($\mathcal{D}$)~\cite{jacot2018neural, chizat2018global}, we assume that ${\bm \sigma}_{\tilde{\mathcal{U}}^g_i} = {\bm \sigma}_{\tilde{\mathcal{U}}_i}$ and ${\bm \sigma}_{\tilde{\mathcal{U}}^g} = {\bm \sigma}_{\tilde{\mathcal{U}}}$. Thus, $\forall \ k \in \left[K\right]$, $$\frac{\left({\bm \mu}^k_{\tilde{\mathcal{U}}^g_i} - {\bm \mu}^k_{\tilde{\mathcal{U}}^g}\right)^2}{\left({\bm \sigma}^k_{\tilde{\mathcal{U}}^g_i}\right)^2 + \left({\bm \sigma}^k_{\tilde{\mathcal{U}}^g}\right)^2} \le \frac{\left({\bm \mu}^k_{\tilde{\mathcal{U}}_i} - {\bm \mu}^k_{\tilde{\mathcal{U}}}\right)^2}{\left({\bm \sigma}^k_{\tilde{\mathcal{U}}_i}\right)^2 + \left({\bm \sigma}^k_{\tilde{\mathcal{U}}}\right)^2}.$$ As this inequality is satisfied in all dimensions including the dimension where the maximum value exists, so for the Fisher discriminant ratio, we have 
    $$F1\left(\tilde{\mathcal{U}}^g_i, \tilde{\mathcal{U}}^g\right) = \max_k\left[\frac{\left({\bm \mu}^k_{\tilde{\mathcal{U}}^g_i} - {\bm \mu}^k_{\tilde{\mathcal{U}}^g}\right)^2}{\left({\bm \sigma}^k_{\tilde{\mathcal{U}}^g_i}\right)^2 + \left({\bm \sigma}^k_{\tilde{\mathcal{U}}^g}\right)^2}\right] \le \max_k\left[\frac{\left({\bm \mu}^k_{\tilde{\mathcal{U}}_i} - {\bm \mu}^k_{\tilde{\mathcal{U}}}\right)^2}{\left({\bm \sigma}^k_{\tilde{\mathcal{U}}_i}\right)^2 + \left({\bm \sigma}^k_{\tilde{\mathcal{U}}}\right)^2}\right] = F1\left(\tilde{\mathcal{U}}_i, \tilde{\mathcal{U}}\right).$$ The smaller the Fisher discriminant ratio is, the less discriminative the two domains are. The less discriminative the two domains are, the smaller $d_{\mathcal{H}}\left(\cdot, \cdot\right)$ is. Thus, finally, we have $$d_{\mathcal{H}}\left(\tilde{\mathcal{U}}^g_i, \tilde{\mathcal{U}}^g\right) \le d_{\mathcal{H}}\left(\tilde{\mathcal{U}}_i, \tilde{\mathcal{U}}\right).$$
\end{proof}

\subsection{Derivations of Corollary 2}
\label{sec:corollary2}

When we focus on the global-to-local knowledge transfer on the \textit{client side}, in the FL scenario, we rewrite \cref{the:1} as

\begin{theorem}
    Consider a virtual global data domain $\mathcal{D}$ and a local data domain $\mathcal{D}_i$. Let $\mathcal{D} = \langle \mathcal{U}, c^* \rangle$ and $\mathcal{D}_i = \langle \mathcal{U}_i, c^* \rangle$, where $\mathcal{U} \subseteq \mathcal{X}$ and $\mathcal{U}_i \subseteq \mathcal{X}$. Given a feature extraction function $\mathcal{F}: \mathcal{X} \mapsto \mathcal{Z}$ that shared between $\mathcal{D}$ and $\mathcal{D}_i$, a random labeled sample of size $m$ generated by applying $\mathcal{F}$ to a random sample from $\mathcal{U}$ labeled according to $c^*$, then for every $h\in \mathcal{H}$, with probability at least $1-\delta$: 
    $$ 
    \mathcal{L}_{\mathcal{D}_i}\left(h\right) \le \mathcal{L}_{\hat{\mathcal{D}}}\left(h\right) + \sqrt{\frac{4}{m} \left(d\log\frac{2em}{d} + \log\frac{4}{\delta}\right)} + d_{\mathcal{H}}\left(\tilde{\mathcal{U}}, \tilde{\mathcal{U}}_i\right) + \lambda_i, 
    $$
    where $\tilde{\mathcal{U}}_i$ and $\tilde{\mathcal{U}}$ are the induced distributions of $\mathcal{U}_i$ and $\mathcal{U}$ under $\mathcal{F}$, respectively. $\tilde{\mathcal{U}}_i \subseteq \mathcal{Z}$ and $\tilde{\mathcal{U}} \subseteq \mathcal{Z}$. $\lambda_i := \min_{h} \mathcal{L}_{\mathcal{D}}\left(h\right) + \mathcal{L}_{\mathcal{D}_i}\left(h\right)$ denotes an oracle performance. \label{the:3}
\end{theorem}

\begin{corollary}
    Let $\mathcal{D}_i$, $\mathcal{D}$, $\mathcal{F}^g$, and $\lambda_i$ defined as in \cref{corollary:mr1}. Given a translation transformation function $\prbm: \mathcal{Z} \mapsto \mathcal{Z}$ that shared between $\mathcal{D}_i$ and \textit{virtual} $\mathcal{D}$, a random labeled sample of size $m$ generated by applying $\mathcal{F}'$ to a random sample from $\mathcal{U}_i$ labeled according to $c^*$, $\mathcal{F}' = \prbm \circ \mathcal{F}^g : \mathcal{X} \mapsto \mathcal{Z}$, then for every $h'\in \mathcal{H}$, with probability at least $1-\delta$: 
    $$ 
    \mathcal{L}_{\mathcal{D}_i}\left(h'\right) \le \mathcal{L}_{\hat{\mathcal{D}}}\left(h'\right) + \sqrt{\frac{4}{m} \left(d\log\frac{2em}{d} + \log\frac{4}{\delta}\right)} + d_{\mathcal{H}}\left(\tilde{\mathcal{U}}', \tilde{\mathcal{U}}'_i\right) + \lambda_i, 
    $$
    where $d_{\mathcal{H}}\left(\tilde{\mathcal{U}}', \tilde{\mathcal{U}}'_i\right) = d_{\mathcal{H}}\left(\tilde{\mathcal{U}}^g, \tilde{\mathcal{U}}^g_i\right) \le d_{\mathcal{H}}\left(\tilde{\mathcal{U}}, \tilde{\mathcal{U}}_i\right) = d_{\mathcal{H}}\left(\tilde{\mathcal{U}_i}, \tilde{\mathcal{U}}\right)$.  $\tilde{\mathcal{U}}'$ and $\tilde{\mathcal{U}}'_i$ are the induced distributions of $\mathcal{U}$ and $\mathcal{U}_i$ under $\mathcal{F}'$, respectively.  \label{corollary:pm1}
\end{corollary}

\begin{proof}
    \prbm is a translation transformation with parameters $\bar{\bm z}^p_i$, $s.t. \ \forall \ {\bm x}_i \in \mathcal{U}_i, {\bm z}_i = {\bm z}^g_i + \bar{\bm z}^p_i$, where ${\bm z}_i = \mathcal{F}'\left({\bm x}_i\right) \in \tilde{\mathcal{U}}'_i$ and ${\bm z}^g_i = \mathcal{F}^g\left({\bm x}_i\right) \in \tilde{\mathcal{U}}^g_i$. In other words, $\forall \ {\bm z}^g_i \in \tilde{\mathcal{U}}^g_i, \exists! \ {\bm z}_i \in \tilde{\mathcal{U}}'_i.$ Therefore, we have ${\rm Pr}_{\tilde{\mathcal{U}}^g_i}\left[\{{\bm z} \in \mathcal{Z}\}\right] = {\rm Pr}_{\tilde{\mathcal{U}}'_i}\left[\{{\bm z} \in \mathcal{Z}\}\right]$ and the same applies to the pair of $\tilde{\mathcal{U}}^g$ and $\tilde{\mathcal{U}}'$, \ie, ${\rm Pr}_{\tilde{\mathcal{U}}^g}\left[\{{\bm z} \in \mathcal{Z}\}\right] = {\rm Pr}_{\tilde{\mathcal{U}}'}\left[\{{\bm z} \in \mathcal{Z}\}\right]$. Then the subtraction of the probability on each side is also equal, \ie, $${\rm Pr}_{\tilde{\mathcal{U}}^g_i}\left[\{{\bm z} \in \mathcal{Z}\}\right] - {\rm Pr}_{\tilde{\mathcal{U}}^g}\left[\{{\bm z} \in \mathcal{Z}\}\right] = {\rm Pr}_{\tilde{\mathcal{U}}'_i}\left[\{{\bm z} \in \mathcal{Z}\}\right] - {\rm Pr}_{\tilde{\mathcal{U}}'}\left[\{{\bm z} \in \mathcal{Z}\}\right].$$  $\forall \ h' \in \mathcal{H}, h^g = h' \circ \prbm \in \mathcal{H}$, so $\forall \ {\bm z}^a \in \mathcal{Z}$ if $h^g\left({\bm z}^a\right) = 1$, then $h'\left({\bm z}^b\right) = 1$, where $\ {\bm z}^b = {\bm z}^a + \bar{\bm z}^p_i$.  Therefore, we have $${\rm Pr}_{\tilde{\mathcal{U}}^g_i}\left[\mathcal{Z}_{h^g}\right] - {\rm Pr}_{\tilde{\mathcal{U}}^g}\left[\mathcal{Z}_{h^g}\right] = {\rm Pr}_{\tilde{\mathcal{U}}'_i}\left[\mathcal{Z}_{h'}\right] - {\rm Pr}_{\tilde{\mathcal{U}}'}\left[\mathcal{Z}_{h'}\right],$$
    where $\mathcal{Z}_{h^g} = \{{\bm z} \in \mathcal{Z} : h^g\left({\bm z}\right) = 1\}, h^g \in \mathcal{H}$ and $\mathcal{Z}_{h'} = \{{\bm z} \in \mathcal{Z} : h'\left({\bm z}\right) = 1\}, h' \in \mathcal{H}$. According to \cref{def}, we have
    $$
    \begin{aligned}
        d_{\mathcal{H}}\left(\tilde{\mathcal{U}}', \tilde{\mathcal{U}}'_i\right) 
        &= 2\sup_{h' \in \mathcal{H}} \left| {\rm Pr}_{\tilde{\mathcal{U}}'_i}\left[\mathcal{Z}_{h'}\right] - {\rm Pr}_{\tilde{\mathcal{U}}'}\left[\mathcal{Z}_{h'}\right] \right| \\
        &= 2\sup_{h^g \in \mathcal{H}} \left| {\rm Pr}_{\tilde{\mathcal{U}}^g_i}\left[\mathcal{Z}_{h^g}\right] - {\rm Pr}_{\tilde{\mathcal{U}}^g}\left[\mathcal{Z}_{h^g}\right] \right|\\
        &= d_{\mathcal{H}}\left(\tilde{\mathcal{U}}^g, \tilde{\mathcal{U}}^g_i\right) \\
        &\le d_{\mathcal{H}}\left(\tilde{\mathcal{U}}, \tilde{\mathcal{U}}_i\right). 
    \end{aligned}
    $$
\end{proof}

\section{Detailed Settings}
\label{sec:setting}

\subsection{Implementation Details}

We create the datasets for each client using six  public datasets: Fashion-MNIST (FMNIST)\footnote{\url{https://pytorch.org/vision/stable/datasets.html\#fmnist}}, Cifar100\footnote{\url{https://pytorch.org/vision/stable/datasets.html\#cifar}}, Tiny-ImageNet\footnote{\url{http://cs231n.stanford.edu/tiny-imagenet-200.zip}} (100K images with 200 labels) and AG News\footnote{\url{https://pytorch.org/text/stable/datasets.html\#ag-news}} (a news classification dataset with four labels, more than 30K samples per label). The MDL is calculated through the public code\footnote{\url{https://github.com/willwhitney/reprieve}}. We run all experiments on a machine with two Intel Xeon Gold 6140 CPUs (36 cores), 128G memory, eight NVIDIA 2080 Ti GPUs, and CentOS 7.8. 

\subsection{Hyperparameters of \mpm}

For hyperparameter tuning, we use grid search to find optimal hyperparameters, including $\kappa$ and $\mu$. Specifically, grid search is performed in the following search space:
\begin{itemize}
    \item $\kappa$: {$0$, $0.001$, $0.01$, $0.1$, $1$, $5$, $10$, $20$, $50$, $100$, $200$, $500$}
    \item $\mu$: {$0$, $0.1$, $0.2$, $0.3$, $0.4$, $0.5$, $0.6$, $0.7$, $0.8$, $0.9$, $1.0$}
\end{itemize}

In this paper, we set $\kappa=50, \mu=1.0$ for the 4-layer CNN, $\kappa=1, \mu=0.1$ for the ResNet-18, and $\kappa=0.1, \mu=1.0$ for the fastText. We only set different values for the hyperparameters $\kappa$ and $\mu$ on different model architectures but use identical settings for one architecture on all datasets. Different models exhibit diverse capabilities in both feature extraction and classification. Given that our proposed $\texttt{DBE}$ operates by integrating itself into a specific model, it is crucial to tune the parameters $\kappa$ and $\mu$ to adapt to the feature extraction and classification abilities of different models. 

As for the \textit{criteria for hyperparameter tuning}, $\kappa$ and $\mu$ require different tunning methods according to their functions. Specifically, $\mu$ is a momentum introduced along with the widely-used moving average technology in approximating statistics, so for the model architectures that originally contain statistics collection operations (\eg, the batch normalization layers in ResNet-18) one can set a relatively small value by tuning $\mu$ from 0 to 1 with a reasonable step size. For other model architectures, one can set a relatively large value for $\mu$ by tuning it from 1 to 0. The parameter $\kappa$ is utilized to regulate the magnitude of the MSE loss in \mr. However, different architectures generate feature representations with varying magnitudes, leading to differences in the magnitude of the MSE loss. Thus, we tune $\kappa$ by aligning the magnitude of the MSE loss with the other loss term. 

\section{Additional Experiments}
\label{sec:addexp}

\subsection{Convergence}
\label{sec:converge}

\begin{figure}[ht]
	\centering
	\includegraphics[width=0.8\linewidth]{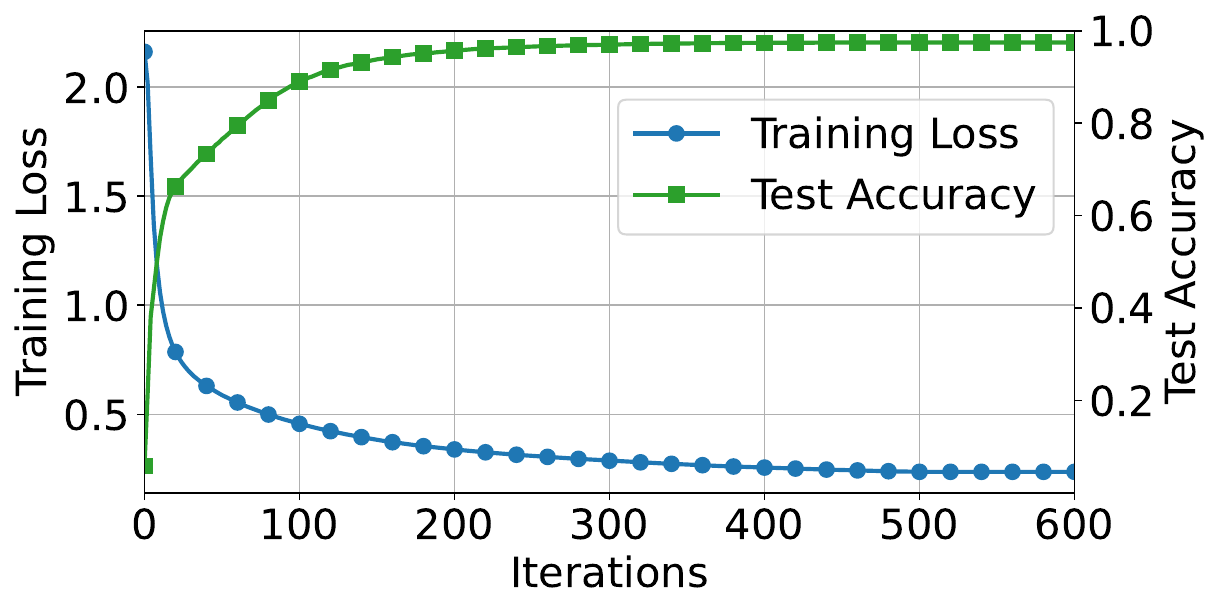}
	\caption{The training loss and test accuracy curve of FedAvg+\mpm on FMNIST dataset using the 4-layer CNN in the practical setting. }
	\label{fig:converge}
\end{figure}

Recall that our objective is 
\begin{equation}
    \min_{{\bm \theta}_1, \ldots, {\bm \theta}_N} \ \mathbb{E}_{i \in [N]}[\mathcal{L}_{\mathcal{D}_i}({\bm \theta}_i)],  \label{eq:pFL1} 
\end{equation}
and its empirical version is $\min_{{\bm \theta}_1, \ldots, {\bm \theta}_N} \sum^{N}_{i=1} \frac{n_i}{n} \mathcal{L}_{\hat{\mathcal{D}}_i}({\bm \theta}_i)$. Here, we visualize the value of $\sum^{N}_{i=1} \frac{n_i}{n} \mathcal{L}_{\hat{\mathcal{D}}_i}({\bm \theta}_i)$ and the corresponding test accuracy during the FL process. \Cref{fig:converge} shows the convergence of FedAvg+\mpm and its stable training procedure. 
Besides, we also report the total iterations required for convergence  on Tiny-ImageNet using ResNet-18 in \Cref{tab:misc_supply}. Based on the findings from \Cref{tab:misc_supply}, we observe that the utilization of \mpm can yield a substantial reduction from 230 to 107 (more than 50\%) in the total number of communication iterations needed for convergence, as compared to the original requirements of FedAvg.

\subsection{Model-Splitting in ResNet-18}

In the main body, we have shown that \mpm improves the per-layer MDL and accuracy of FedAvg no matter how we split the 4-layer CNN. In \Cref{tab:perlayer_resnet_tiny}, we report the per-layer MDL and accuracy when we consider model splitting in ResNet-18, a model deeper than the 4-layer CNN. No matter at which layer, we split ResNet-18 to form a feature extractor and a classifier, \mpm can also reduce MDL and improve accuracy, showing its general applicability.

\begin{table}[h]
  \centering
  \caption{The MDL (bits, $\downarrow$) of layer-wise representations, test accuracy (\%, $\uparrow$), and the number of trainable parameters ($\downarrow$) in \prbm when adding \mpm to FedAvg on Tiny-ImageNet using ResNet-18 in the practical setting. The ``B'', ``CONV'', ``POOL'', and ``FC'' means the ``block'', ``convolution block'', ``average pool layer'', and ``fully connected layer'' in ResNet-18~\cite{he2016deep}, respectively. }
  \resizebox{\linewidth}{!}{
    \begin{tabular}{l|ccccccc|c|c}
    \toprule
    \multirow{2}{*}{\textbf{Metrics}} & \multicolumn{7}{c|}{\textbf{MDL}} & \multirow{2}{*}{\textbf{Accuracy}} & \multirow{2}{*}{\textbf{Param.}} \\
    \cmidrule{2-8}
     & CONV$\rightarrow$B1 & B1$\rightarrow$B2 & B2$\rightarrow$B3 & B3$\rightarrow$B4 & B4$\rightarrow$POOL & POOL$\rightarrow$FC & Logits & & \\
    \midrule
    Original (FedAvg) & 4557 & 4198 & 3598 & 3501 & 3445 & 3560 & 3679 & 19.45 & 0 \\
    \midrule
    CONV$\rightarrow$\mpm$\rightarrow$B1 & \underline{4332} & 4050 & 3528 & 3407 & 3292 & 3347 & 3493 & 19.96 & 16384 \\
    B1$\rightarrow$\mpm$\rightarrow$B2 & 4527 & \underline{4072} & 3568 & 3456 & 3361 & 3451 & 3560 & 19.50 & 16384 \\
    B2$\rightarrow$\mpm$\rightarrow$B3 & 4442 & 4091 & \underline{3575} & 3474 & 3326 & 3411 & 3520 & 19.55 & 8192 \\
    B3$\rightarrow$\mpm$\rightarrow$B4 & 4447 & 4073 & 3511 & \underline{3414} & 3259 & 3346 & 3467 & 20.72 & 4096 \\
    B4$\rightarrow$\mpm$\rightarrow$POOL & 4424 & 4030 & 3391 & 3304 & \underline{3284} & 3511 & 3612 & 39.99 & 2048 \\
    POOL$\rightarrow$\mpm$\rightarrow$FC & 4432 & 4035 & 3359 & 3298 & 3209 & \underline{3454} & 3594 & 42.98 & 512 \\
    \bottomrule
    \end{tabular}}
  \label{tab:perlayer_resnet_tiny}
\end{table}

\subsection{Distinguishable Representations}

\begin{figure}[h]
	\centering
	\includegraphics[width=0.5\linewidth]{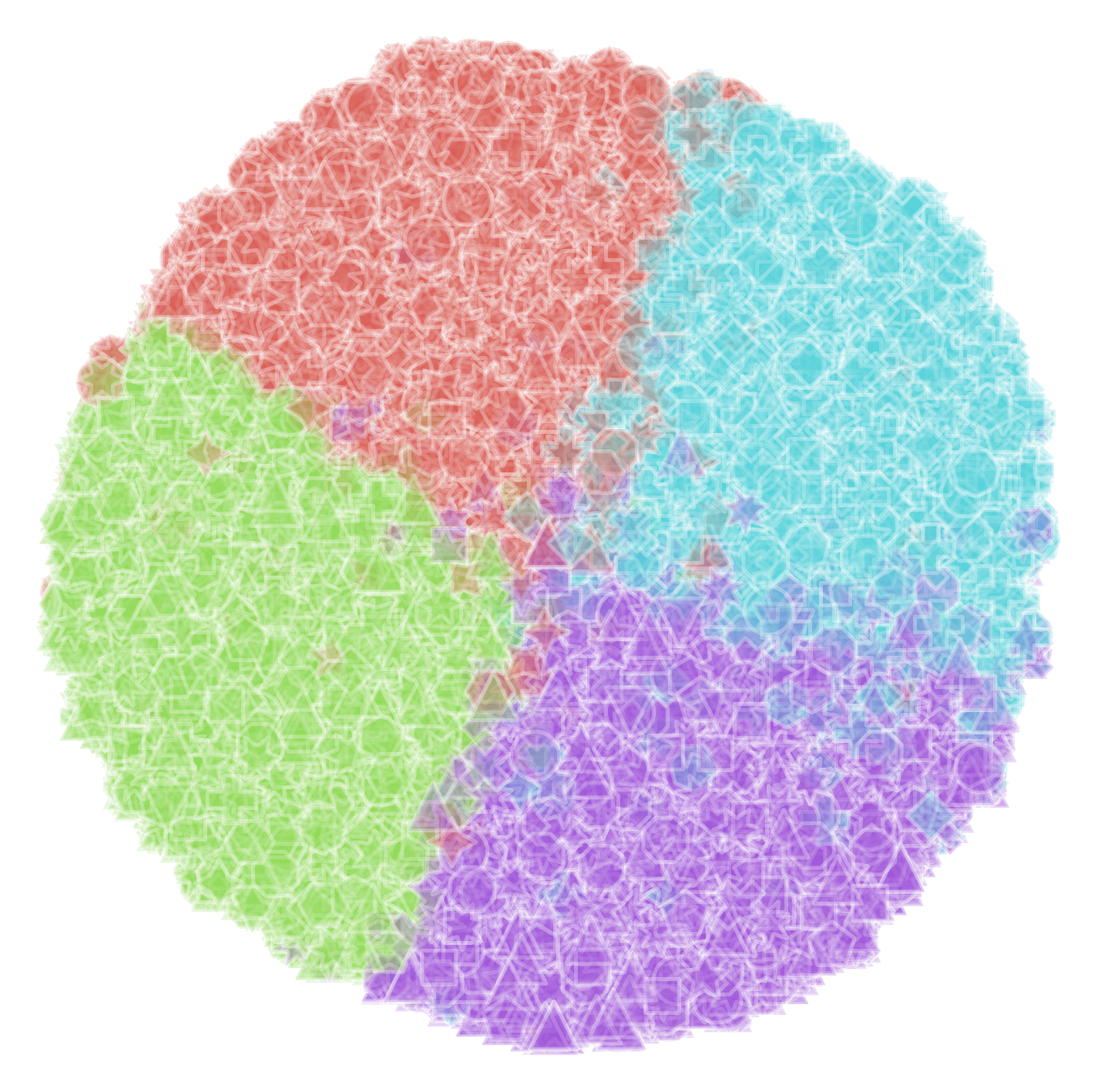}
	\caption{t-SNE visualization for the representations extracted by the global feature extractor on AG News (four labels) in FedAvg+$\texttt{DBE}$. We use \textit{color} and \textit{shape} to distinguish \textit{labels} and \textit{clients}, respectively.}
 \label{fig:distinguishable}
\end{figure}

As our primary goal is to demonstrate the elimination of representation bias rather than improving discrimination in Figure 3 (main body), we present the t-SNE visualization for our largest dataset in experiments, Tiny-ImageNet (200 labels). Given that the 200 labels are distributed around the chromatic circle, adjacent labels are assigned similar colors, resulting in Figure 3 (main body) being indistinguishable by the label. Using a dataset AG News with only four labels for t-SNE visualization can clearly show that the representations extracted by the global feature extractor are distinguishable in \Cref{fig:distinguishable}. 

\subsection{A Practical Scenario with New Participants}

To simulate a practical scenario with new clients joining for future FL, we perform method-specific local training for 10 epochs on new participants for warming up after their local models are initialized by the learned global model (or client models in FedFomo). Since FedAvg, Per-FedAvg, and FedBABU do not generate personalized models during the FL process, we fine-tune the entire global model on new clients for them to obtain test accuracy. Specifically, using Cifar100 and 4-layer CNN, we conduct FL on 80 old clients ($\rho=0.5$ or $\rho=0.1$) and evaluate accuracy on 20 new joining clients after warming up. We utilize the data distribution depicted in \Cref{fig:distribution-100}. According to \Cref{tab:misc_supply}, FedAvg shows excellent generalization ability with fine-tuning. However, \mpm can still improve FedAvg by up to \textbf{\textcolor{green_}{+6.68}} with more stable performance for different $\rho$. 

\begin{table}[h]
  \centering
  \caption{The total iterations for convergence and the averaged test accuracy (\%, $\uparrow$) of pFL methods. }
  \resizebox{!}{!}{
    \begin{tabular}{l|r|cc|ccc}
    \toprule
    \textbf{Items} & \multicolumn{1}{c|}{\textbf{Iterations}} & \multicolumn{2}{c|}{\textbf{New Participants}} & \multicolumn{3}{c}{\textbf{Local Epochs}} \\
    \midrule
    &  & $\rho=0.5$ & $\rho=0.1$ & 1 & 5 & 10 \\
    \midrule
    Per-FedAvg~\cite{NEURIPS2020_24389bfe} & 34 & 48.66 & 48.36 & 95.10 & 93.92 & 93.91 \\
    pFedMe~\cite{t2020personalized} & 113 & 41.20 & 38.39 & 97.25 & 97.44 & 97.32 \\
    Ditto~\cite{li2021ditto} & 27 & 36.57 & 45.06 & 97.47 & 97.67 & 97.64 \\
    FedPer~\cite{arivazhagan2019federated} & 43 & 39.86 & 42.39 & 97.44 & 97.50 & 97.54 \\
    FedRep~\cite{collins2021exploiting} & 115 & 38.75 & 35.09 & 97.56 & 97.55 & 97.55 \\
    FedRoD~\cite{chen2021bridging} & 50 & 50.10 & 51.73 &97.52 & 97.49 & 97.35 \\
    FedBABU~\cite{ohfedbabu} & 513 & 48.60 & 42.29 & 97.46 & 97.57 & 97.65 \\
    APFL~\cite{deng2020adaptive} & 57 & 38.19 & 45.16 &97.25 & 97.31 & 97.34 \\
    FedFomo~\cite{zhang2020personalized} & 71 & 27.50 & 27.47 & 97.21 & 97.17 & 97.22 \\
    APPLE~\cite{ijcai2022p301} & 45 & --- & --- & 97.06 & 97.07 & 97.01 \\
    \midrule
    FedAvg  & 230 & 52.52 & 49.44 & 85.85 & 85.96 & 85.53 \\
    FedAvg+\mpm & 107 & \textbf{57.62} & \textbf{56.12} & \textbf{97.69} & \textbf{97.75} & \textbf{97.78} \\
    \bottomrule
    \end{tabular}}
  \label{tab:misc_supply}
\end{table}

\subsection{Large Local Epochs}

We also conduct experiments with more local epochs in each iteration on FMNIST using the 4-layer CNN, as shown in \Cref{tab:misc_supply}. All the pFL methods perform similarly with the results for one local epoch, except for Per-FedAvg, which degenerates around 1.18 in accuracy (\%). 

\begin{table}[h]
  \centering
  \caption{The test accuracy (\%) on the HAR dataset. }
  \resizebox{!}{!}{
    \begin{tabular}{l|c}
    \toprule
    Methods & Accuracy \\
    \midrule
    FedAvg & 87.20$\pm$0.27 \\
    SCAFFOLD & 91.34$\pm$0.43 \\
    FedProx & 88.34$\pm$0.24 \\
    MOON & 89.86$\pm$0.18 \\
    FedGen & 90.82$\pm$0.21 \\
    Per-FedAvg & 77.12$\pm$0.17 \\
    pFedMe & 91.57$\pm$0.12 \\
    Ditto & 91.53$\pm$0.09 \\
    FedPer & 75.58$\pm$0.13 \\
    FedRep & 80.44$\pm$0.42 \\
    FedRoD & 89.91$\pm$0.23 \\
    FedBABU & 87.12$\pm$0.31 \\
    APFL & 92.18$\pm$0.51 \\
    FedFomo & 63.39$\pm$0.48 \\
    APPLE & 86.46$\pm$0.35 \\
    \midrule
    FedAvg+$\texttt{DBE}$ & \textbf{94.53$\pm$0.26} \\
    \bottomrule
    \end{tabular}}
  \label{tab:app}
\end{table}

\subsection{Real-World Application}

We also evaluate the performance of our \mpm in a real-world application. Specifically, we apply \mpm to the Internet-of-Things (IoT) scenario on a popular Human Activity Recognition (HAR) dataset~\cite{anguita2012human} with the HAR-CNN~\cite{zeng2014convolutional} model. HAR contains the sensor signal data collected from 30 users who perform six activities (WALKING, WALKING\_UPSTAIRS, WALKING\_DOWNSTAIRS, SITTING, STANDING, LAYING) wearing a smartphone on the waist. We show the results in \Cref{tab:app}, where FedAvg+\mpm still achieves superior performance. 

\section{Broader Impacts}
\label{sec:impact}

The representation bias and representation degeneration naturally exist in FL under statistically heterogeneous scenarios, which are derived from the inherently separated local data domains on individual clients. In the main body, we show the general applicability of our proposed \mpm to representative FL methods. More than that, \mpm can also be applied to other practical fields, such as the Internet of Things (IoT)~\cite{nguyen2021federated, ghimire2022recent, hu2020personalized} and digital health~\cite{chen2020fedhealth, chen2022metafed}. Furthermore, introducing the view of knowledge transfer into FL sheds light on this field. 

\section{Limitations}
\label{sec:limit}

Although FL comes along for privacy-preserving and collaborative learning, it still suffers from privacy leakage issues with malicious clients~\cite{zhang2022fldetector, cao2021provably} or under attacks~\cite{fang2020local, luo2021feature}. We design \mpm based on FL to improve generalization and personalization abilities, and we only modify the local training procedure without affecting the downloading, uploading, and aggregation processes. Thus, the \mpm-equipped FL methods still suffer from the originally existing privacy issues like the original version of these FL methods when attacks happen. It requires future work to devise specific methods for privacy-preserving enhancement. 

\section{Data Distribution Visualization}
\label{sec:disvis}

We illustrate the data distributions (including training and test data) in our experiments here. 

	% \subfigure[FedAvg (B).]{\includegraphics[width=0.16\linewidth]{fig/study_rep_new_tsne_Tiny-imagenet-0.1_cnn_FedAvg-save_rep_test__01.png}\label{fig:fed_before}}

\begin{figure*}[ht]
	\centering
	\hfill
    \subfigure[FMNIST]{\includegraphics[width=0.32\linewidth]{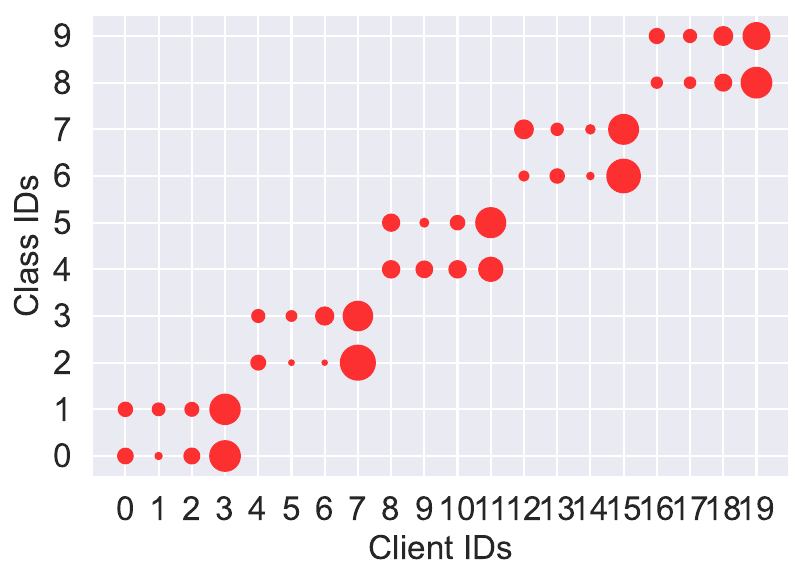}}
	\hfill
    \subfigure[Cifar100]{\includegraphics[width=0.32\linewidth]{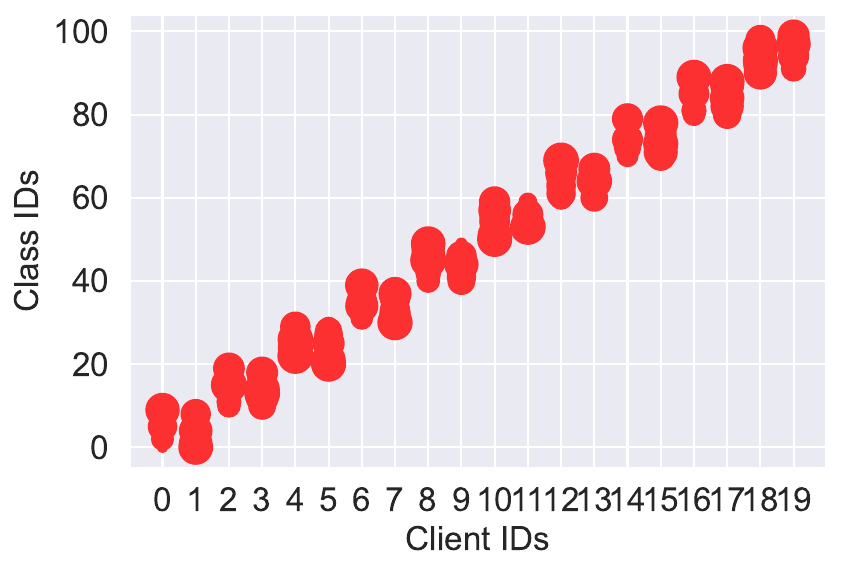}}
    \hfill
    \subfigure[Tiny-ImageNet]{\includegraphics[width=0.32\linewidth]{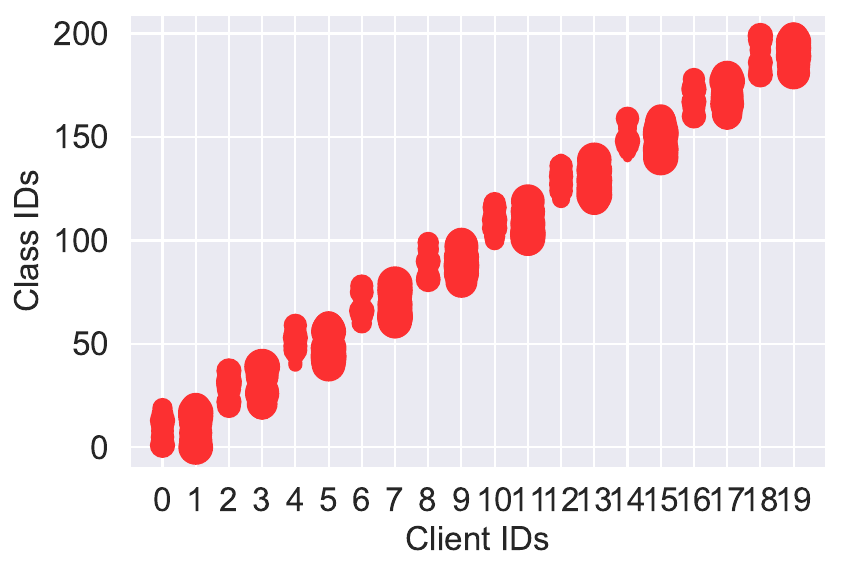}}
    \hfill
	\hfill
	\caption{The data distributions of all clients on FMNIST, Cifar100, and Tiny-ImageNet, respectively, in the pathological settings. The size of a circle represents the number of samples. }
	\label{fig:distribution-pathological}
\end{figure*}

\begin{figure*}[ht]
	\centering
    \hfill
    \subfigure[FMNIST]{\includegraphics[width=0.32\linewidth]{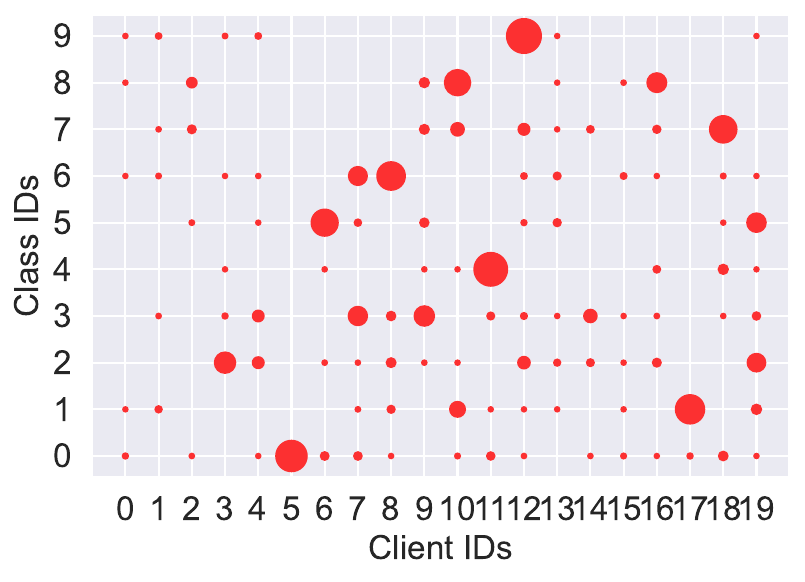}}
	\hfill
    \subfigure[Cifar100]{\includegraphics[width=0.32\linewidth]{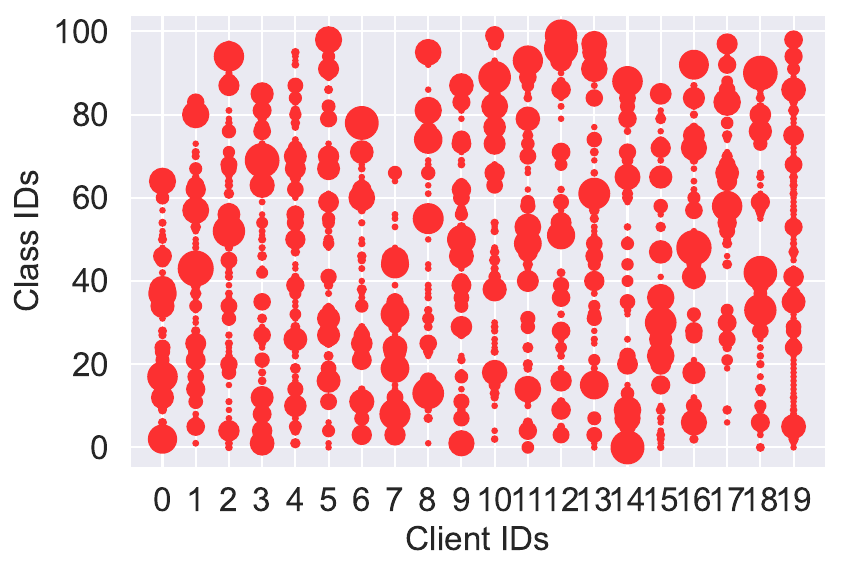}}
    \hfill
    \subfigure[Tiny-ImageNet]{\includegraphics[width=0.32\linewidth]{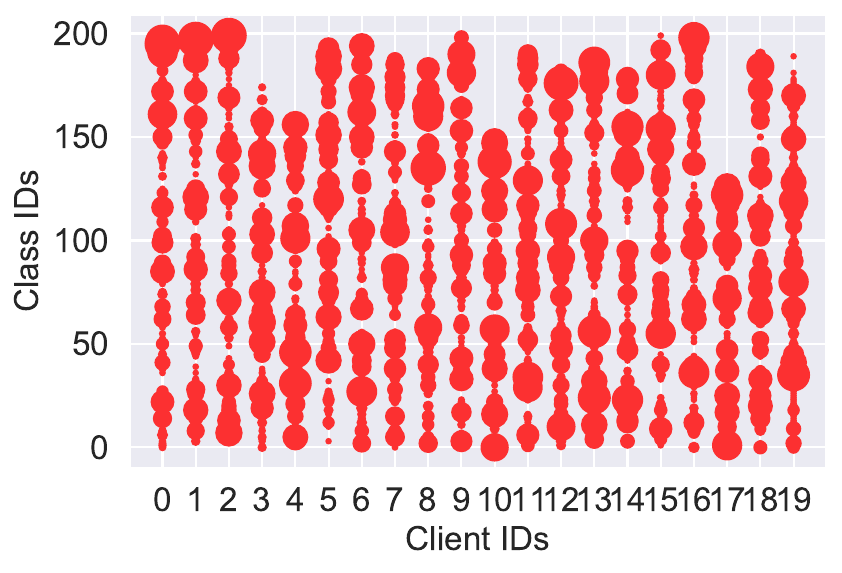}}
    \hfill
	\hfill
	\caption{The data distributions of all clients on FMNIST, Cifar100, and Tiny-ImageNet, respectively, in the practical settings ($\beta=0.1$). The size of a circle represents the number of samples. }
	\label{fig:distribution-practical}
\end{figure*}

\begin{figure*}[ht]
	\centering
	\hfill
    \subfigure[$\beta=0.01$]{\includegraphics[width=0.32\linewidth]{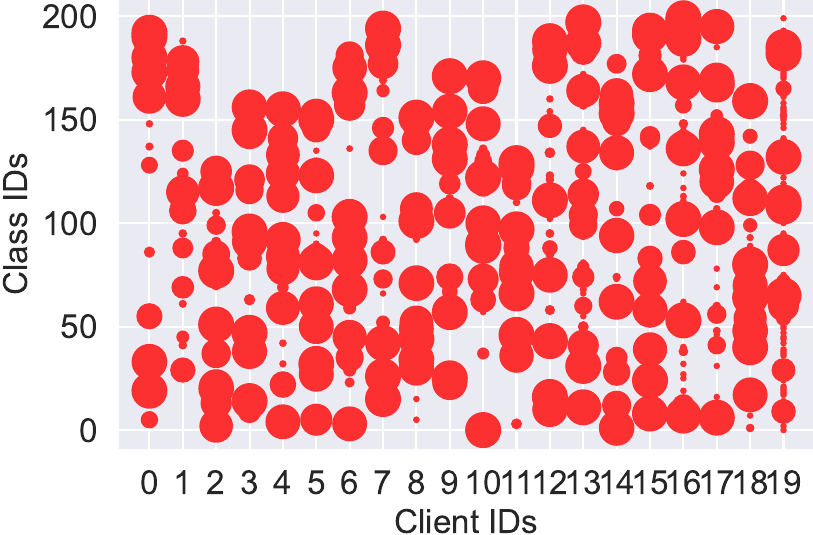}}
	\hfill
    \subfigure[$\beta=0.5$]{\includegraphics[width=0.32\linewidth]{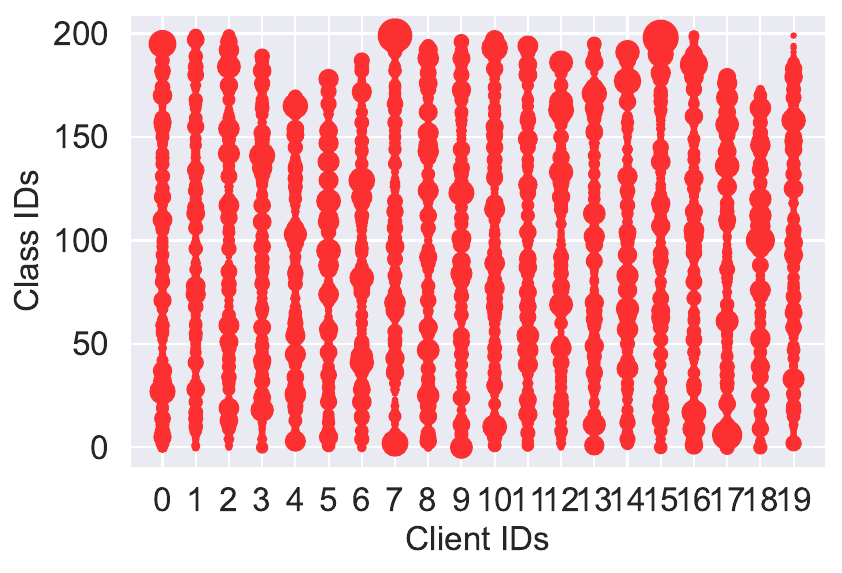}}
    \hfill
    \subfigure[$\beta=5$]{\includegraphics[width=0.32\linewidth]{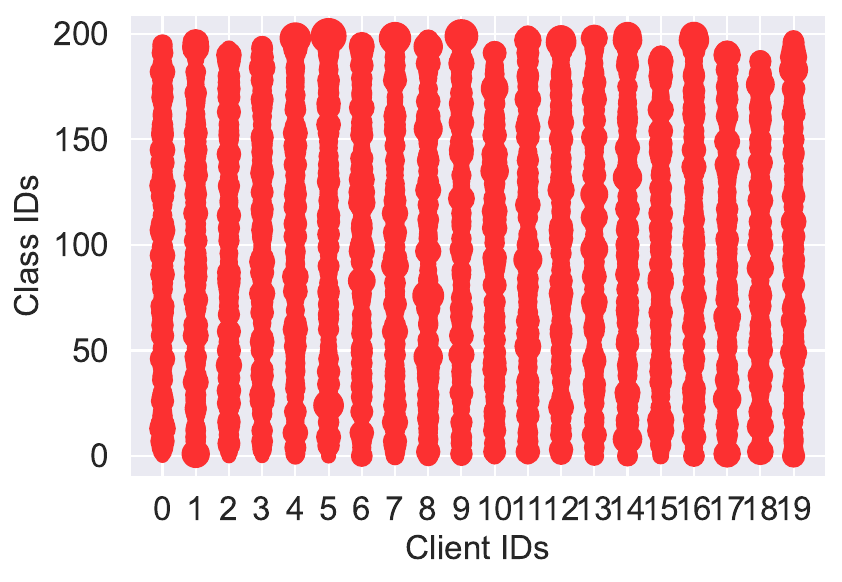}}
	\hfill
	\hfill
	\caption{The data distribution on all clients on Tiny-ImageNet in three additional practical settings. The size of a circle represents the number of samples. The degree of heterogeneity decreases as $\beta$ in $Dir(\beta)$ increases.}
	\label{fig:distribution-hetero}
\end{figure*}

\begin{figure*}[ht]
	\centering
	\includegraphics[width=\textwidth]{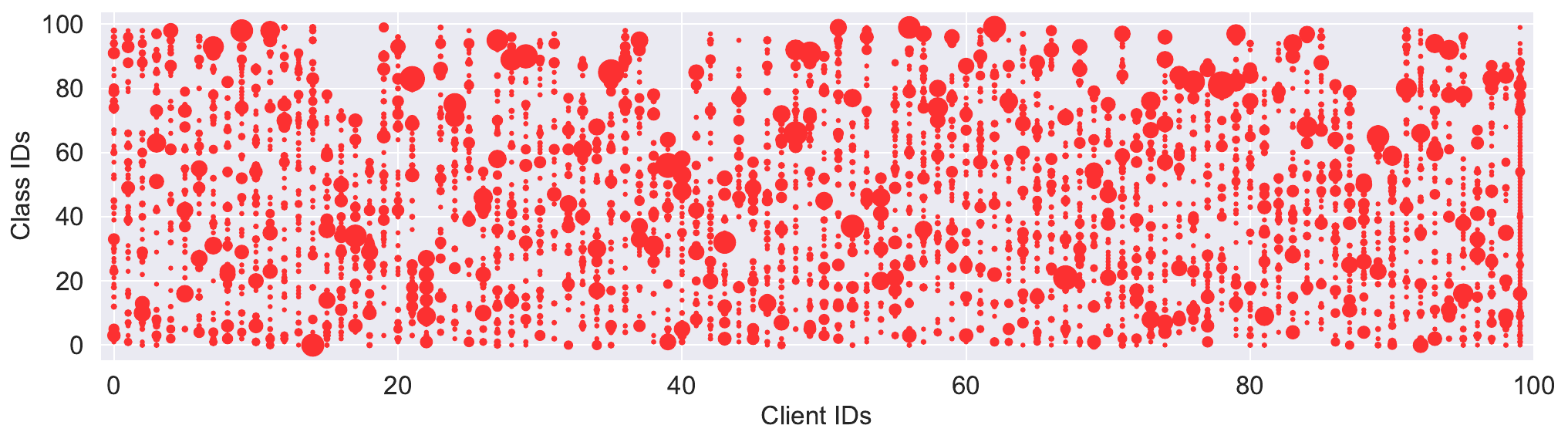}
	\caption{The data distributions of all clients on Cifar100 in the practical setting ($\beta=0.1$) with 100 clients, respectively. The size of a circle represents the number of samples. }
	\label{fig:distribution-100}
\end{figure*}

\end{document}